\documentclass[10pt,journal,compsoc]{IEEEtran}
\usepackage{graphicx}
\usepackage{subfigure}
\usepackage{amsmath}
\usepackage{amssymb}
\usepackage{booktabs}
\usepackage{hhline}
\usepackage{multirow}
\usepackage{multicol}
\usepackage{microtype}      
\usepackage{xcolor}         
\usepackage{hhline}
\usepackage{bbm}
\usepackage{boldline}
\usepackage{graphicx}
\usepackage{amssymb}
\usepackage{pifont}
\usepackage{soul}
\usepackage[accsupp]{axessibility}
\usepackage{float} 
\newcommand{\cmark}{\ding{51}}%
\usepackage{hyperref}
\hypersetup{
    colorlinks=true,
    linkcolor=blue,
    filecolor=magenta,      
    urlcolor=cyan,
}
\usepackage[font=footnotesize]{caption}
\usepackage[capitalize]{cleveref}
\crefname{section}{Sec.}{Secs.}
\Crefname{section}{Section}{Sections}
\Crefname{table}{Table}{Tables}
\crefname{table}{Tab.}{Tabs.}

\ifCLASSOPTIONcompsoc
  \usepackage[nocompress]{cite}
\else
  \usepackage{cite}
\fi
\ifCLASSINFOpdf
\else
\fi
\newcommand{\Skip}[1]
{
}
\newcommand{\JP}[1]
{
    \textcolor{blue}{\bfseries{{JP: #1}}}
}

\begin{document}
%
\title{Task-Oriented Channel Attention for\\Fine-Grained Few-Shot Classification}
%
%
%
%

\author{SuBeen Lee, WonJun Moon, Hyun Seok Seong, and Jae-Pil Heo,~\IEEEmembership{Member,~IEEE,}
\IEEEcompsocitemizethanks{
\IEEEcompsocthanksitem SuBeen Lee, WonJun Moon, and Hyun Seok Seong are with the Department of Artificial Intelligence, Sungkyunkwan University, South Korea, 16419.\protect\\
E-mail: \{leesb7426, wjun0830, gustjrdl95\}@skku.edu
\IEEEcompsocthanksitem Jae-Pil Heo is with the Department of Computer Science and Engineering and Department of Artificial Intelligence, Sungkyunkwan University, South Korea, 16419.\protect\\
E-mail: jaepilheo@skku.edu
}
\thanks{Corresponding author: Jae-Pil Heo}}

%
%

\markboth{Journal of \LaTeX\ Class Files,~Vol.~14, No.~8, August~2015}%
{Shell \MakeLowercase{\textit{et al.}}: Bare Demo of IEEEtran.cls for Computer Society Journals}
%



\IEEEtitleabstractindextext{%

\begin{abstract}
The difficulty of the fine-grained image classification mainly comes from a shared overall appearance across classes. Thus, recognizing discriminative details, such as eyes and beaks for birds, is a key in the task. However, this is particularly challenging when training data is limited. To address this, we propose Task Discrepancy Maximization (TDM), a task-oriented channel attention method tailored for fine-grained few-shot classification with two novel modules Support Attention Module (SAM) and Query Attention Module (QAM). SAM highlights channels encoding class-wise discriminative features, while QAM assigns higher weights to object-relevant channels of the query. Based on these submodules, TDM produces task-adaptive features by focusing on channels encoding class-discriminative details and possessed by the query at the same time, for accurate class-sensitive similarity measure between support and query instances. While TDM influences high-level feature maps by task-adaptive calibration of channel-wise importance, we further introduce Instance Attention Module (IAM) operating in intermediate layers of feature extractors to instance-wisely highlight object-relevant channels, by extending QAM. The merits of TDM and IAM and their complementary benefits are experimentally validated in fine-grained few-shot classification tasks. Moreover, IAM is also shown to be effective in coarse-grained and cross-domain few-shot classifications.
\end{abstract}

}

\maketitle

\IEEEdisplaynontitleabstractindextext

%
\IEEEpeerreviewmaketitle

\section{Introduction}
\label{sec:intro}
\IEEEPARstart{D}{eep} learning has made great strides in various vision tasks, even achieving remarkable performance beyond humans in many downstream tasks~\cite{deng2009imagenet, he2016deep}.
However, such performance is achieved under the presence of numerous labeled images, which require huge labeling costs.
In other words, the performance can be significantly degraded if the number of labeled images is insufficient~\cite{matchnet, maml, closer}.
Therefore, such limited condition from a shortage of labeled images and the high cost of labeling promotes the growth of few-shot classification~\cite{maml, protonet, matchnet}, which is to train a model highly adaptable to novel classes.
To achieve this goal, the training of the few-shot classification is mainly based on the episodic learning strategy, where each episode comprises a few sampled categories from the dataset.
In addition, images of each class are split into a support set and a query set for the training and evaluation, respectively.

The metric-based learning is a mainstream of the few-shot classification~\cite{matchnet, protonet, khrulkov2020hyperbolic,sung2018learning}.
These methods learn a deep representation with a predefined or online-trained metric,
and the inference for a query is performed based on the distances among support and query sets under such metric.
However, since the feature extractor is only trained with base classes, the feature maps for novel classes computed by the learned extractor, hardly form a tight cluster~\cite{zhang2021prototype, roady2020difsim}.
To alleviate this, recent methods utilize primitive knowledge~\cite{zhang2021prototype, li2020boosting} or propose task-dynamic feature alignment strategies~\cite{kang2021relational, xu2021learning, ctx, dsn, frn, feat, hou2019cross}. 
Among two strategies, task-dynamic feature alignment approaches are being spotlighted and can be further divided into two main streams; spatial alignment and channel alignment. 
The spatial alignment methods~\cite{hou2019cross, ctx, xu2021learning, kang2021relational, frn, frn} aim to resolve the spatial mismatch between key features on the feature maps of different instances.
On the other hand, since the semantic feature maps of novel classes are not optimized for each episode, the channel alignment methods try to adapt those feature maps to the target classification task by considering the composition of the episode.

Although aforementioned alignment methods accomplish huge improvements on the coarse-grained few-shot classification task, they provide insignificant gains for fine-grained datasets.
This is mainly because they only focus on exploit channel or spatial information which may not be discriminative for the episode.
Indeed, localizing discriminative details is important in fine-grained classification, since categories are highly likely to share similar overall appearances~\cite{ge2019weakly, liu2020filtration, ding2019selective, zheng2019looking}.
Therefore, distinct clues for each category, which have only subtle differences from other categories, should also be captured for fine-grained few-shot classification.

In this context, we introduce a novel module, Task Discrepancy Maximization (TDM), that localizes discriminative regions by weighting channels per class. 
TDM highlights the channels representing discriminative regions and restrains the contributions of other channels based on a class-wise channel weight vector.
Specifically, TDM is composed of two components: Support Attention Module (SAM) and Query Attention Module (QAM).
Given a support set, SAM produces a support weight vector per class that presents high activations on discriminative channels. 
On the other hand, QAM is fed with the query set to output a query weight vector per instance, where such query weight vector highlights the object-relevant channels.
To compute these weight vectors, the relation between each feature map and the corresponding channel-wisely average pooled feature is considered.
Since the channel-wisely average pooled feature map has the spatial information of the object~\cite{cbam, li2020spatial},
channels are highly likely to represent salient regions when they are similar to spatially averaged feature maps.
By combining two weight vectors computed from our sub-modules, a task-specific weight vector is finally defined.
Consequently, the task-specific weight vector is utilized to produce task-adaptive feature maps which replace the original feature maps.


Although TDM is a tailored module for the fine-grained few-shot classification task, its performance can be highly dependent on the quality of given feature maps produced by the feature extractor since the TDM is designed to work with high-level feature maps. Therefore, we further introduce IAM, Instance Attention Module as an extended version of QAM, to implement our main idea even for the feature extraction. Unlike QAM, IAM operates in the intermediate layers of the feature extractor and computes a channel weight vector per instance to enhance the quality of the feature representation like existing attention methods~\cite{cbam, senet, selfattention}.
Since IAM induces the feature extractor to focus on instance-wise informative channels, the resulting feature map contains more object-relevant information and less background information. As mentioned, the IAM is designed to complement the TDM in the feature extraction stage, interestingly, however, it also helps to boost the performances of the general few-shot classification task.
\Skip{
            Although TDM is a well-tailored module for fine-grained few-shot classification, the quality of its output can be dependent on the feature map.
            Therefore, we further introduce Instance Attention Module (IAM), an extended version of QAM.
            Unlike QAM, IAM operates in the intermediate layer of the feature extractor and computes a channel weight vector per instance to enhance the quality of the feature representation like existing attention methods~\cite{cbam, senet, selfattention}.
            Since IAM induces the feature extractor to focus on instance-wise information channels, the feature map contains more object-relevant information and less background-relevant information.
            Note that, IAM is initially designed to complement TDM, but it can be also adopted in the general few-shot classification.
}

Our main contributions are summarized as follows:
\begin{itemize}
\item We propose a novel feature alignment method, TDM, to define the class-wise channel importance based on identifying class-discriminative and query-relevant channels, tailored for the fine-grained few-shot classification task.
\item We further extend QAM to introduce IAM by reflecting the concepts of TDM to the feature extractor, which not only complements TDM in the fine-grained tasks but also benefits for more general scenarios including coarse-grained and cross-domain few-shot classification.
\item We experimentally validate the high applicability of proposed TDM and IAM to the prior few-shot classification models and strength of them by achieving new state-of-the-art performances in standard benchmarks.
\end{itemize}

\Skip{
            Our main contributions are summarized as follows:
            \begin{itemize}
            \item We propose a novel feature alignment method, TDM, to define the class-wise channel importance, for fine-grained few-shot classification.
            \item Our proposed TDM is highly applicable to prior metric-based fine-grained few-shot classification models.
            \item We extend QAM to introduce IAM which not only complements TDM in fine-grained datasets but is also utilized in more general scenarios like coarse-grained and cross-domain.
            \item Our experiments with popular few-shot classification models verify the strength of TDM and IAM with new state-of-the-art performances.
            \end{itemize}
}
\section{Related Work}
\label{sec:rel}
\subsection{Few-Shot Classification}
There are two main streams in the few-shot image classification research, the optimization- and metric-based approaches. At early stage, MAML introduced the concept of optimization-based methods where it learns good initial conditions for adaptation to the novel tasks. Then, Meta-LSTM~\cite{ravi2016optimization} used an LSTM-based meta-learner for general initial point and effective fine-tuning.
Moreover, MetaOptNet~\cite{metaoptnet} provides a differentiation process for end-to-end learning by utilizing convex base learners.
Although these optimization-based methods show promising results, they need online updates for novel classes.

On the other hand, the metric-based methods aim to learn deep representations by adopting a predefined~\cite{matchnet, protonet, khrulkov2020hyperbolic} or online-trained metric~\cite{sung2018learning}.
Its concept is introduced in MatchNet~\cite{matchnet} which infers categories of the query set by the cosine similarity.
ProtoNet~\cite{protonet} further employs a mean feature of each class as a prototype and utilizes them for computing the distance between a query and each class.
Instead of the predefined metrics, RelationNet~\cite{sung2018learning} exploits a learnable distance metric.

As aforementioned, metric-based methods generally try to reduce distances among instances belonging to the same category. TDM is an applicable module for those metric-based methods to boost thier performance. Specifically, TDM enables the distances to be measured based on adaptive channel weights where it identifies and emphasizes discriminative channels dynamically, while prior techniques treat all the channels equally.

\subsection{Feature Alignment Methods}

In the metric-based classification, feature alignment methods are developed for classification-friendly distance computations. These feature alignment approaches can be classified into spatial and channel alignment methods.
The spatial alignment methods~\cite{hou2019cross, ctx, xu2021learning, kang2021relational, frn, deepemd} aim to align the features of the support and query sets to match object regions. For example, CAN~\cite{hou2019cross} computes a correlation map for each pair of the classes and query feature maps to emphasize the common regions where the object likely to exist.
CTX~\cite{ctx} measures a coarse spatial correspondence between the query instance and the support set by the attention~\cite{bahdanau2014neural}, then it produces a query-aligned prototype per each class based on the correspondence.
FRN~\cite{frn} reconstructs the feature maps of the support set in accordance with the feature map of the query instance by exploiting a closed-form solution of the ridge regression.

On the other hand, the channel alignment methods~\cite{xu2021learning, kang2021relational, feat, dsn} alter feature maps so that the novel classes are well distinguished.
Specifically, FEAT manipulates the feature maps of support sets to increase the distance among classes by utilizing the transformer~\cite{lin2017structured, vaswani2017attention}.
DMF~\cite{xu2021learning} aligns each feature map of the query instances by the dynamic meta-filter produced in the consideration of the support and query pair.
And, RENet~\cite{kang2021relational} transforms feature maps of the support and query pair with self- and cross-correlation which capture the structural patterns of each image and encode semantically relevant contents, respectively.


Although TDM is basically a channel alignment method, 
unlike existing methods that typically consider a pairwise relationship between the support set for each category and the query instance, TDM utilizes the entire task to adapt the feature maps.

\subsection{Attention Modules}
In various downstream tasks, attention modules yield remarkable performance gain~\cite{tdm, lap, move, transvpr, HP}.
The existing attention methods can be divided into spatial attentions~\cite{selfattention, vit, swin} and channel attentions~\cite{senet, cbam}.
Specifically, to resolve poor scaling properties of CNN, SA~\cite{selfattention} proposes a spatial attention module to capture long-range dependencies in an image by representing each grid of the feature map with other regions and itself based on similarity.
Based on the success of SA, ViT~\cite{vit} proposes the network architecture, only comprised of fully-connected layers and multi-head attention, and shows a powerful performance of spatial attention.

On the other hand, SENet~\cite{senet} proposes a channel attention module that produces channel weights highlighting more informative channels from spatially averaged features via a simple fully-connected block.
CBAM~\cite{cbam} is another notable channel attention module and it used not only averaged feature but also max-pooled features.

Among the two attention approaches, TDM and IAM belong to channel attention methods. 
However, unlike the existing attention methods that only consider the information of an instance to produce a result of attention, TDM computes a channel weight vector for each class and query instance with the entire task.
It is to estimate the category of the query instance by focusing on query-relevant features among the distinct characteristics of each class.
Furthermore, IAM induces the feature map of each instance to consist of object-relevant features for minimizing the impact of background in TDM.
Therefore, we claim that our modules are specialized in the few-shot classification.
\section{Preliminary}
\label{sec:pre}

\subsection{Problem Formulation}\label{problem_formulation}
As a standard formulation of the few-shot classification problem,
we are given two datasets: meta-train set $D_{base}=\left\{\left(x_i,y_i\right),y_i\in{C_{base}}\right\}$ for training a model and meta-test set $D_{novel}=\left\{\left(x_i,y_i\right),y_i\in{C_{novel}}\right\}$ for evaluating a learned model. 
$C_{base}$ and $C_{novel}$ indicate base classes and novel classes, respectively, where they do not overlap (i.e., ${C_{base}}\cap {C_{novel}}=\phi$). 
Generally, training and testing of few-shot classification are composed of episodes. 
Each episode consists of randomly sampled $N$ classes and each class is composed of $K$ labeled images and $U$ unlabeled images, \textit{i.e.}, $N$-way $K$-shot episode. 
The labeled images are called the support set $S=\left\{\left(x_j, y_j\right)\right\}^{{N}\times{K}}_{j=1}$, and the unlabeled images are named the query set $Q=\left\{\left(x_j, y_j\right)\right\}_{j=1}^{{N}\times{U}}$, while two sets are disjoint (i.e., ${S}\cap{Q}=\phi$). 
The support and query sets are utilized for learning and validation, respectively.
Commonly, the category of the query instance is predicted by utilizing feature maps of the support and query instances.
If we define $x^S_{i,j}$ as $j$-th instance of $i$-th class in the support set and $x^Q$ as the query instance, their corresponding feature maps $F^S_{i,j}$ and $F^Q$ are expressed as follows:
\begin{equation}
    \begin{split}
        \label{feature_maps}
        F^S_{i,j}=g_{\theta}(x^S_{i,j})
        \\
        F^Q=g_{\theta}(x^Q),
    \end{split}
\end{equation}
where $g_\theta$ is the feature extractor parameterized by $\theta$. 
The shape of each feature map is $\mathbb{R}^{{C}\times{H}\times{W}}$ where $C,H,$ and $W$ denote the number of channels, height, and width, respectively. 

\subsection{Motivation}
\begin{figure}[t!]
    \centering
    \includegraphics[width=\columnwidth]{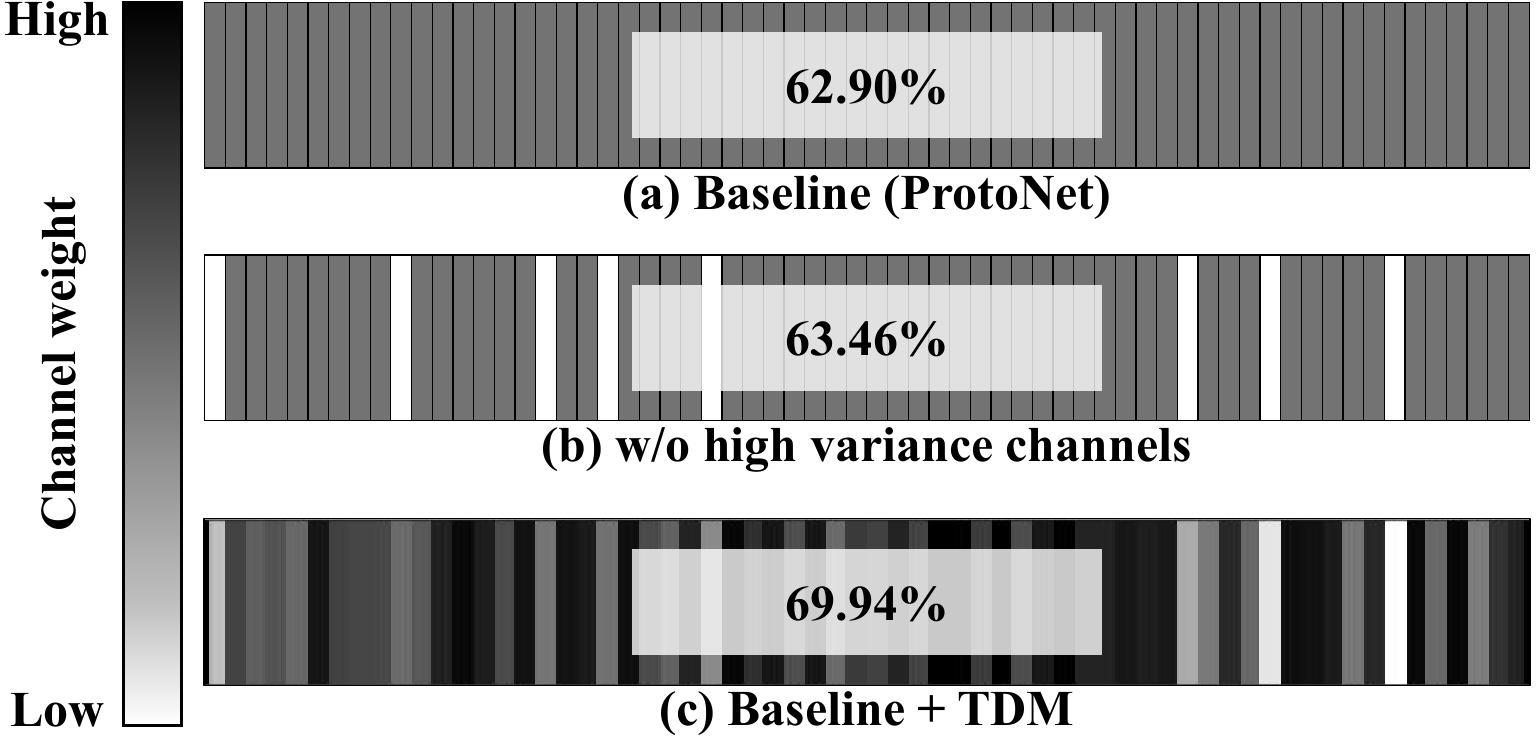}
    \caption{
Effect of the channel weight in the CUB dataset. 
Each column of the sub-figure represents the channel weight, and the numbers in boxes are average classification accuracies of the 5-way 1-shot scenario which are evaluated with 10,000 randomly sampled episodes from novel classes.
(a) Baseline equally treats channels of feature maps regardless of the channel-wise variance within a class ($V_{i,c}$ defined in \cref{variance_within}). 
In a such case, channels possessing high variance within a class are highly likely to disturb the precise estimation of the category
-- intuitively, the instances of the same class having similar features at a channel lead to a low channel variance for the corresponding channel.
(b) Therefore, we can get improvement by just removing channels with high $V_{i,c}$ for each episode in the evaluation phase (we compute the $V_{i,c}$ with support- and query-set, and eliminate the top 12.5\% channels with high $V_{i,c}$).
However, in an episode of fine-grained datasets, even if feature maps of categories possess low $V_{i,c}$ in all channels, those feature maps may not be optimized to the episode.
This is because categories share similar features, i.e., feathers and wings in CUB datasets.
(c) Therefore, in fine-grained datasets, we should focus on whether each channel reflects distinct characteristics.
TDM produces per-class channel weight based on the discriminative power of channels for each class in the episode.
    }
    \label{fig:motivation}
\end{figure}
\begin{figure*}[t]
    \centering
    \includegraphics[width=1.\textwidth]{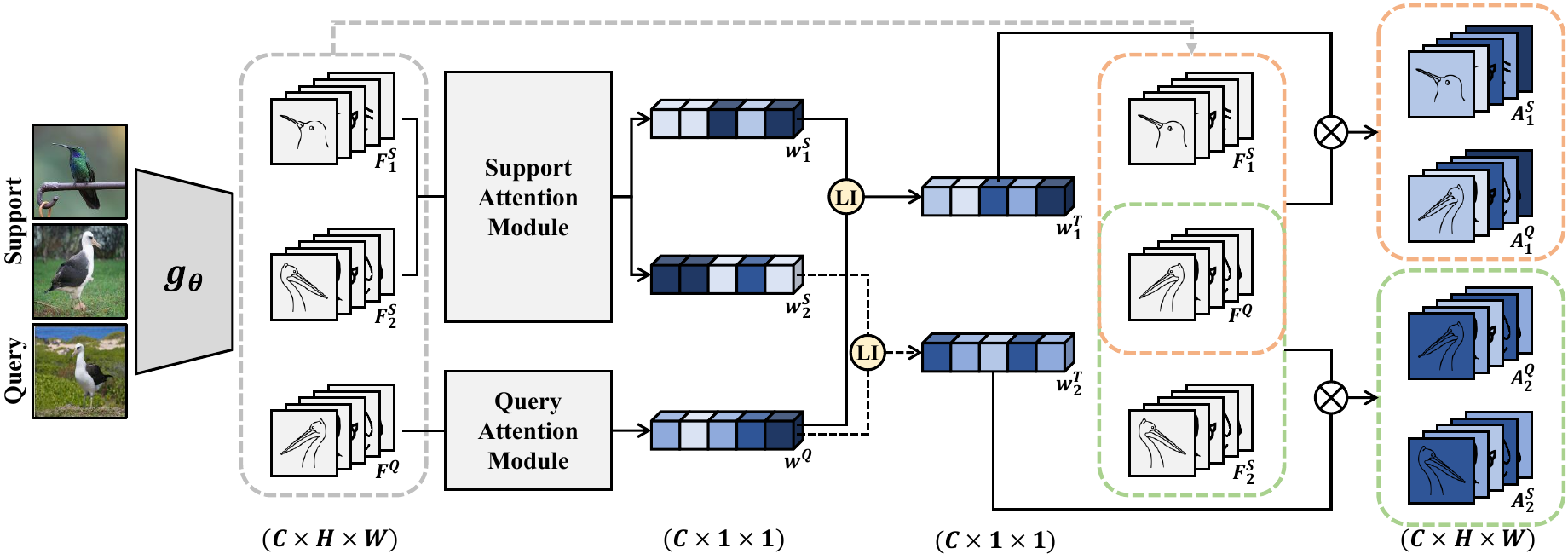}
    \caption{
    Overview of Task Discrepancy Maximization. 
    TDM consists of two sub-modules and each of them takes feature maps $F$ to generate channel weight vector $w$.
    Support attention module (SAM) receives feature maps of the support instances as input and estimates discriminative channels for each category.
    Then, for each $i$-th class, it produces a support weight vector $w^S_i$ where the vector holds high values in those channels.  
    On the other hand, the query attention module (QAM) takes a feature map of the query instance and discovers the object-relevant channels of the query.
    Then, a query weight vector $w^Q$ from the QAM emphasizes those channels with high values. 
    These weight vectors from two sub-modules are combined by linear interpolation to define a task weight vector $w^T_i$ for each $i$-th category.
    Finally, task-adaptive feature maps $A$, which concentrate on the discriminative regions, are obtained by multiplying the task weight vector to the original feature maps.
    }
    \label{fig:TDM}
\end{figure*}
In metric-based few-shot learning~\cite{protonet, matchnet}, the classification is generally performed based on the distances.
Suppose that such distances are defined for $C$-dimensional vectors $s \in \mathbb{R}^C$ computed by channel-wise spatial average of the feature map $F$ as follows:
\begin{equation}
    \begin{split}
        \label{scalar}
        s_{c}&=\frac{1}{H \times W}\sum^{H}_{h=1}\sum^{W}_{w=1}f_{c,h,w},
    \end{split}
\end{equation}
where $f_{c,h,w}$ the value spatially located at $(h,w)$ in $c$-th channel of $F$ which is the feature map of an instance. 
Then, we compute the average value for $c$-th dimension of support- and query-set which belong to the $i$-th class, as follows:
\begin{equation}
    \begin{split}
        \label{bar_scalar}
        \bar{s}_{i,c}&=\frac{1}{J_i}\sum^{J_i}_{j=1}s_{i,j,c},
    \end{split}
\end{equation}
where $J_i$ is the number of instances that belongs to the $i$-th class and $s_{i,j,c}$ denotes $s_c$ of $j$-th instance in $i$-th class.
Based on those, we define the channel-wise variance $V_{i,c}$ of $c$-th channel within $i$-th class as follows:
\begin{equation}
    \begin{split}
        \label{variance_within}
        V_{i,c}=\frac{1}{J_i}\sum^{J_i}_{j=1} {\left( s_{i,j,c} - \bar{s}_{i,c}\right)^2}.
    \end{split}
\end{equation}

Since the values at the same dimension are compared for distance computation among vectors, the dimensions with small variance less contribute to the distance.
Thus, metric-based few-shot classification methods try to reduce $V_{i,c}$ in the training phase.
However, even though the categories used in the training phase form low $V_{i,c}$, the same is not guaranteed for the novel classes in the validation phase~\cite{feat, zhang2021prototype, roady2020difsim}.

Therefore, as described in \cref{fig:motivation} (a), it is not a proper solution that utilizes all channels equally.
This is because there are channels with high $V_{i,c}$ in novel classes since the feature extractor is trained with base classes, as aforementioned.
Thus, as described in \cref{fig:motivation} (b), it is effective to utilize only channels with low $V_{i,c}$ by eliminating channels with high $V_{i,c}$ in novel classes, just as base classes which consist of channels with low $V_{i,c}$.

However, in fine-grained datasets, categories belong to the same super-class and share common features, then, even if channels possess low $V_{i,c}$, they may not contain distinct information from other classes.
Accordingly, we should consider whether the information of each channel is discriminative with other classes as described in \cref{fig:motivation} (c).
To achieve it, we introduce our two novel channel attention modules in \cref{sec:met}.
Further, as described in \cref{sec:gen}, we extend one of them to capture the instance-descriptive information.

\section{Task Discrepancy Maximization : \\Alignment method for fine-grained datasets}
\label{sec:met}
\begin{figure}[t!]
    \centering
    \includegraphics[width=\columnwidth]{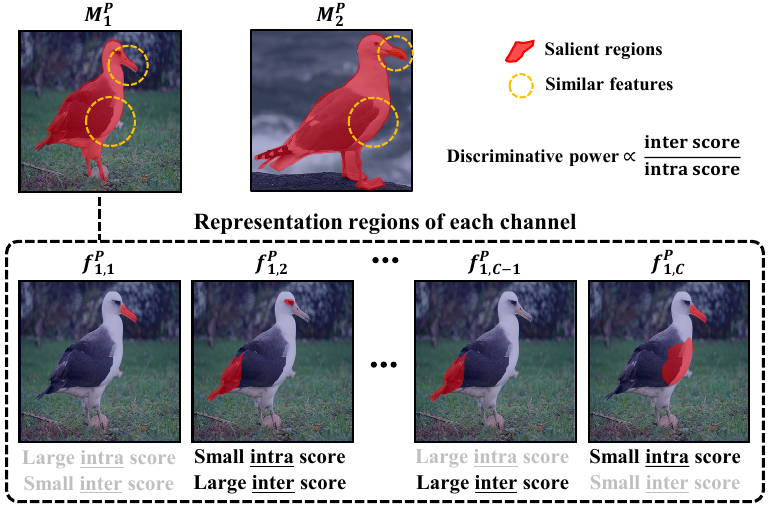}
    \caption{
    Relation between channel-wise representativeness scores and discriminative power.
    Suppose that categories share two similar features (breast feathers and beak) and two different features (tails and eyes).
    If a channel encodes various features of an object, the regions covered by the encoded features highly match with the salient region~(i.e., 2nd and C-th channels). In this case, $R^\text{intra}_{i,j}$ for the corresponding channel is small (i.e., $R^\text{intra}_{1,2}$ and $R^\text{intra}_{1,C}$ are small). 
    Likewise, while the $R^\text{intra}_{i,j}$ is related to the regional coverage of $j$-th channel for $i$-th object category, $R^\text{inter}_{i,j}$ represents the discriminative power of $j$-th channel for $i$-th class.
    For example, $R^\text{inter}_{i,C}$ (of $C$-th channel) is small since it only encodes characteristics shared by another bird category (i.e., breast feathers and beak are similar features shared by $1^{\text{st}}$ and $2^{\text{nd}}$ classes). In contrast, $R^\text{inter}_{i,2}$ is large because the second channel encodes highly discriminative features (i.e., tails and eyes).
    As a result, the discriminative channels should be a small $R^\text{intra}_{i,j}$ and a large $R^\text{inter}_{i,j}$.
    }
    \label{fig:Distances}
\end{figure}

The overall architecture of TDM is illustrated in \cref{fig:TDM}. 
Given an episode that consists of the support and query instances, feature maps are first computed by the feature extractor.
Since the feature extractor is trained to find discriminative features for distinguishing base classes~\cite{feat, zhang2021prototype, roady2020difsim}, the feature maps are not optimal for each episode.
To produce optimized feature maps for each episode, TDM transforms the feature maps by computing task-specific weight vectors representing channel-wise discriminative power for a specific task. 
As a result, TDM aims to refine the original feature maps into task-adaptive feature maps focusing on the discriminatory details.
In this section, we describe the components of TDM and their purposes.
First, we define two channel-wise representativeness scores based on the estimated salient regions in ~\cref{distance}.
Then, with these scores, we introduce two sub-modules of the TDM: SAM and QAM in \cref{support_attention_module} and \cref{query_attention_module}, respectively.
Finally, TDM is described in \cref{task_dircrepancy_maximization} with the discussion in \cref{discussion}.

\subsection{Channel-wise Representativeness Scores}\label{distance}
Given feature maps $F^S_{i,j}$ of the support set, for each pair of $i$-th class and $c$-th channel, we define two channel-wise representativeness scores; intra score $R^{\text{intra}}_{i,c}$, and inter score $R^{\text{inter}}_{i,c}$.
Since there may be multiple instances for each category, we utilize a prototype\cite{protonet} as the representative of each class.
The prototype $F^P_i$ for $i$-th class is computed as follows:
\begin{equation}
    \begin{split}
        \label{prototype}
        F^P_i=\frac{1}{K}\sum^{K}_{j=1}F^S_{i,j},
    \end{split}
\end{equation}
where $K$ and $F^S_{i,j}$ are the number of instances for each class and feature map of $j$-th instance for $i$-th class, respectively.
Then, for each prototype, we define a mean spatial feature to represent salient object regions.
When the $c$-th channel of the prototype $F^P_i$ for $i$-th class is indicated as ${f^P_{i,c}}\in{\mathbb{R}^{{H}\times{W}}}$, the corresponding mean spatial feature $M^P_i$ is computed as follows:
\begin{equation}
    \begin{split}
        \label{mean_spatial_feature}
        M^P_i=\frac{1}{C}\sum^{C}_{j=1}f^P_{i,j}.
    \end{split}
\end{equation}
Based on this, we further define the channel-wise representativeness score within a class,  $R^{\text{intra}}_{i,c}$, for $c$-th channel of $i$-th class as follows:
\begin{equation}
    \begin{split}
        \label{in_class_distance}
        R^{\text{intra}}_{i,c} = \frac{1}{H\times W}\parallel f^P_{i,c} - M^P_i \parallel^2.
    \end{split}
\end{equation}
It represents how well the highly activated regions on the $c$-th channel match the class-wise salient areas represented by the mean spatial feature.
On the other hand, the channel-wise representativeness score across classes, $R^{\text{inter}}_{i,c}$, for $c$-th channel of $i$-th class is computed as follows:
\begin{equation}
    \begin{split}
        \label{out_class_distance}
        R^{\text{inter}}_{i,c} = \frac{1}{H \times W}\min_{ 1 \leq j \leq N, j \neq{i} } \parallel f^P_{i,c} - M^P_j \parallel^2.
    \end{split}
\end{equation}
Since the score is large when $f^P_{i,c}$ is different from $M^P_j$, it denotes how much the channel contains the discriminative information of each category. 
Intuitively, a channel is more distinct when it has a small $R^{\text{intra}}_{i,c}$ and a large  $R^{\text{inter}}_{i,c}$ as illustrated in \cref{fig:Distances}.

\subsection{Support Attention Module (SAM)}\label{support_attention_module}
\begin{figure}[H]
    \centering
    \includegraphics[width=\columnwidth]{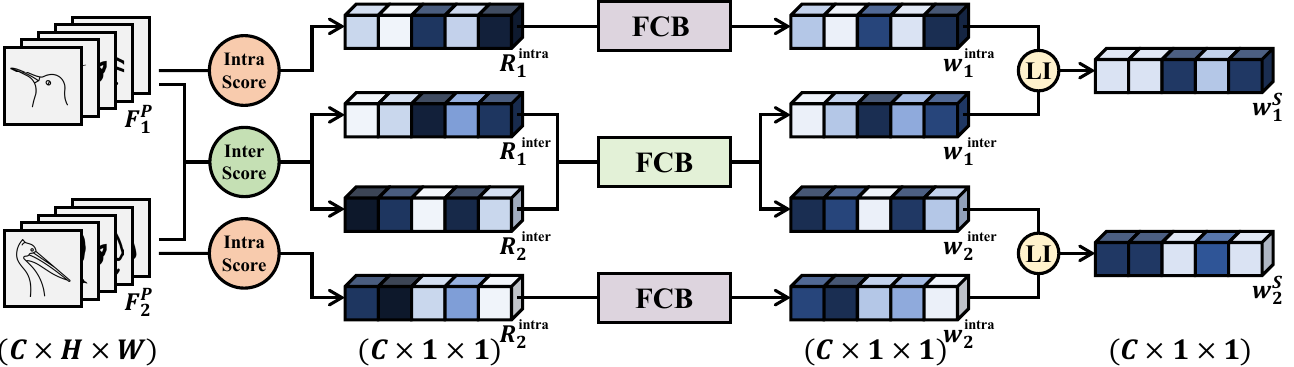}
    \caption{
    Schematic illustration of Support Attention Module.
    }
    \label{fig:SAM}
\end{figure}

Support attention module (SAM) receives the class prototypes as input and computes two channel-wise representativeness scores for each class based on \cref{in_class_distance} and \cref{out_class_distance}.
Then, to reflect the importance of each channel by considering the distribution of those scores, we transform two score vectors, $R^\text{intra}_i$ and $R^\text{inter}_i$, into two weight vectors, $w^{\text{intra}}_i$ and $w^{\text{inter}}_i$ for $i$-th class, as follows:
\begin{equation}
    \begin{split}
        \label{in_out_weight_vector}
        w^{\text{intra}}_i&=b^{\text{intra}}\left(R^{\text{intra}}_{i}\right)
        \\
        w^{\text{inter}}_i&=b^{\text{inter}}\left(R^{\text{inter}}_{i}\right),
    \end{split}
\end{equation}
where $b^{\text{intra}}$ and $b^{\text{inter}}$ denote fully-connected blocks for producing two weight vectors.
The architecture of them are described in \cref{fully_connectec_block}

The support weight vector $w^{S}_i$ for $i$-th class is obtained by linear interpolation of the corresponding two weight vectors, $w^{\text{intra}}_i$ and $w^{\text{inter}}_i$, as follows:
\begin{equation}
    \begin{split}
        \label{support_weight_vector}
        w^{S}_i={\alpha}{w^{\text{intra}}_i}+{\left(1-\alpha\right)}{w^{\text{inter}}_i},\: {\alpha}\in{\left[0,1\right]},
    \end{split}
\end{equation}
where $\alpha$ is a balancing hyperparameter for the support weight vector.
The vector for $i$-th class highlights distinct channels of $i$-th class while suppressing channels that include common information throughout classes in the episode.
Therefore, when the support weight vector $w^{S}_i$ for $i$-th class is multiplied to the feature maps, instances of $i$-th class are gathered, while others become separated from the $i$-th class.

\subsection{Query Attention Module (QAM)}\label{query_attention_module}
\begin{figure}[h]
    \centering
    \includegraphics[width=\columnwidth]{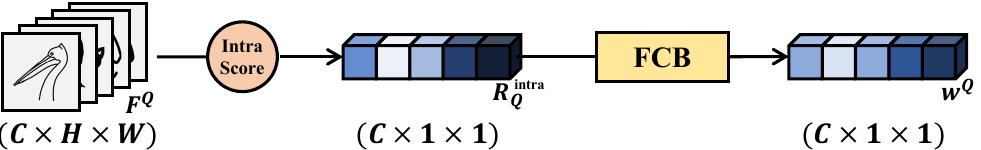}
    \caption{Schematic illustration of Query Attention Module}
    \label{fig:QAM}
\end{figure}
Although the support weight vector $w^S_i$ for $i$-th class is develop to emphasize the distinct channels for $i$-th class, the query instance which belongs to the $i$-th category may not possess those features. Specifically, the emphasized discriminative channels of the support set become useless or they even disturb the distance-based class prediction for the query, if the query does not have such features corresponding to the highlighted channels. This problem motivates us to propose query attention module (QAM) to focus on the channels which are class-discriminative and taken by the query at the same time.
Since we do not have any label information for the query instance unlike the support set, QAM only utilizes the relationship among channels within the query instance.
Specifically, QAM computes the channel-wise representativeness score within the query instance, $R^\text{intra}_{c}$, for $c$-th channel, as follows:
\begin{equation}
    \begin{split}
        \label{qwer}
        R^{\text{intra}}_{c} = \frac{1}{H\times W}\parallel f^{Q}_{c} - M^Q \parallel^2,
    \end{split}
\end{equation}
where $f^Q_{u}$ denotes $c$-th channel of the feature map $F^Q$, and $M^Q$ is the mean spatial feature which is defined by the channel-wise average of $F^Q$.
Then, the query weight vector $w^Q$ is produced by passing the intra score vector $R^\text{intra}$ to the fully-connected block $b^Q$, as follows:
\begin{equation}
    \begin{split}
        \label{query_weight_vector}
        w^Q=b^{\text{Q}}\left(R^{\text{intra}}\right),
    \end{split}
\end{equation}
where the architecture of $b^Q$ is described in \cref{fully_connectec_block}.
The query weight vector emphasizes object-relevant channels of the query instance while restraining other channels. 
Therefore, the query weight vector guides the model to focus on information related to the object of the query.

\subsection{Task Discrepancy Maximization (TDM)}\label{task_dircrepancy_maximization}
\begin{figure}[t!]
    \centering
    \includegraphics[width=\columnwidth]{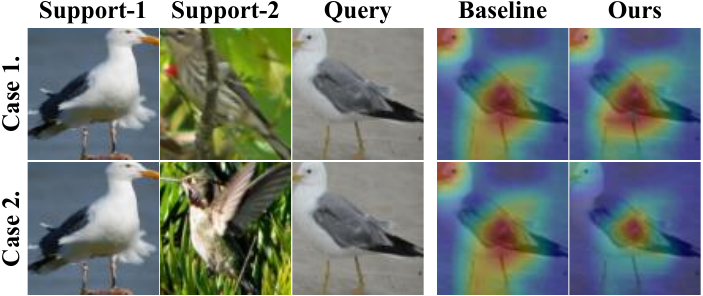}
    \caption{
    2D-aggregated feature activation maps on 2-way 1-shot.
    (Case 1) If beaks and wings are not similar between species, TDM regards both beaks and wings to be discriminative.
    (Case 2) However, when birds have similar beaks, TDM considers only wings as a discriminative part.
    }
    \label{fig:channel_attention_map}
\end{figure}
Since two weight vectors $w^{S}_i$ and $w^{Q}$ respectively produced by the SAM and QAM are complementary in their purposes, we utilize them to define a task weight vector.
Specifically, the task weight vector $w^T_i$ for $i$-th class is defined by linear interpolation of the corresponding support and query weight vectors, $w^S_i$ and $w^Q$, as follows:
\begin{equation}
    \begin{split}
        \label{task_weight_vector}
        w^{T}_i={\beta} {w^{S}_i}+{\left(1-\beta\right)} {w^{Q}},\quad {\beta}\in{\left[0,1\right]},
    \end{split}
\end{equation}
where $\beta$ is a balancing hyperparameter.

Based on the above task weight vectors, the feature maps of all the support and query instances are transformed into task-adaptive feature maps.
Specifically, the feature maps $F^S_{i,j} \in \mathbb{R}^{{C}\times{H}\times{W}}$ of the support instances for $i$-th class are converted to the task-adaptive feature map $A^S_{i,j}$ by its corresponding task weight vector $w^T_i \in \mathbb{R}^{{C}}$, as follows:
\begin{equation}
    A^S_{i,j} = \left[{w^T_{i,1}}{f^S_{i,j,1}},{w^T_{i,2}}{f^S_{i,j,2}}, ... , {w^T_{i,C}}{f^S_{i,j,C}} \right],
\end{equation}
where $w^T_{i,c}$ and $f^S_{i,j,c}$ are a scalar value at $c$-th dimension of $w^T_i$ and the $c$-th channel of $F^S_{i,j}$, respectively.
On the other hand, since the label of the query is not available, it is not possible to specify which task weight vector should be multiplied by the feature map of the query. 
Therefore, we apply all the task weight vectors $w^T$ to the feature map $F^Q\in \mathbb{R}^{{C}\times{H}\times{W}}$ of the query to produce task-adaptive feature maps $A^Q$ about all categories, as follows:
\begin{equation}
    A^Q_{i} = \left[{w^T_{i,1}}{f^Q_{1}},{w^T_{i,2}}{f^Q_{2}}, ... , {w^T_{i,C}}{f^Q_{C}} \right],
\end{equation}
where $i$ indicates class index, and $f^Q_c$ is the $c$-th channel of $F^Q$.
When we are testing the query for $i$-th class, the corresponding adaptive feature map $A^Q_i$ of the query is utilized.

For instance, when TDM is applied to the ProtoNet\cite{protonet}, the probability that the query instance belongs to $i$-th class is computed by the following criteria:
\begin{equation}
    \begin{split}
        \label{Inferring_class}
        p_{\theta}(y=i|x)=\frac{{\exp}(-d(A^P_i,A^Q_i))}{\sum_{j=1}^{N}{\exp}(-d(A^P_j,A^Q_j))},
    \end{split}
\end{equation}
where $d$ is the distance metric, and $A^P_i$ is the prototype computed by the average of the adaptive feature maps of support instances for $i$-th class.

\subsection{Discussion}\label{discussion}
For a general dataset, it is widely known that the feature map, which contains various information about the object, is beneficial~\cite{huang2020self, li2021learning, li2021beyond}.
On the other hand, in a fine-grained dataset, it is advantageous to focus only on the discriminative regions since the categories share a common overall apperance~\cite{ge2019weakly, liu2020filtration, ding2019selective, zheng2019looking}.
Moreover, in fine-grained few-shot classification, the distinct parts of each class may be variable depending on the contents of the episode, unlike general fine-grained classification where the discriminative regions of each category is almost constant.
Thus, dynamically discovering the distinct parts of each class in the episode is the key point in the fine-grained few-shot classification. As described in \cref{fig:channel_attention_map}, the baseline model estimates the category of the query by treating all characteristics equally regardless of the composition of each episode. In contrast, TDM predicts the class of the query by concentrating on discriminative parts which are discovered with consideration for the episode.
This is why TDM is a tailored module for the fine-grained few-shot classification.
\section{Instance Attention Module: \\Generalized Query Attention Module}
\label{sec:gen}
\begin{figure*}[h]
    \centering
    \includegraphics[width=2\columnwidth]{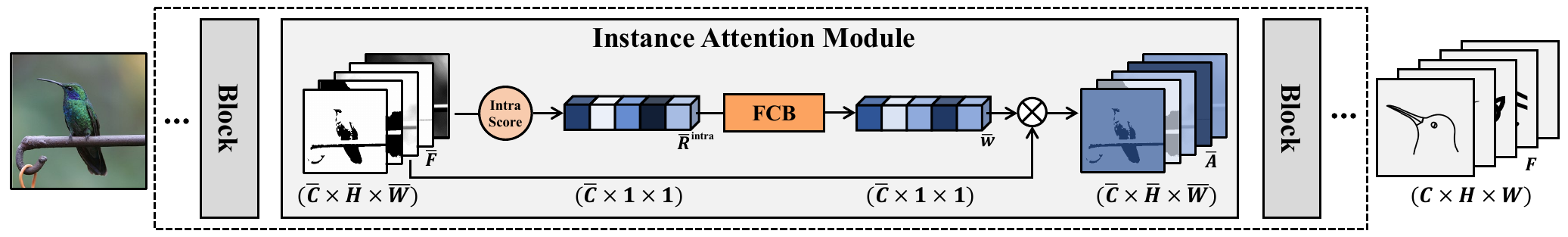}
    \caption{
    Schematic illustration of Instance Attention Module (IAM).
    The box with the dashed line indicates the feature extractor $g_\theta$ in \cref{fig:TDM}. For each instance, IAM receives an intermediate feature map $\overline{F}$ and computes channel-wise representativeness scores $\overline{R}^\text{intra}$ based on the similarity between each channel $\overline{f}_c$ and the salient object region. Then, it produces a channel weight vector $\overline{w} \in \mathbb{R}^{\overline{C}}$ that possesses high values in object-relevant channels of the instance. By scaling each channel of intermediate feature map $\overline{f}_c$ by its corresponding the weight $\overline{w}_c$, an attentive feature map $\overline{A}$ is obtained. Finally, the attention-applied  feature map $\overline{A}$ is passed to the next layer.
    }
    \label{fig:IAM}
\end{figure*}

\Skip{
            Schematic illustration of Instance Attention Module (IAM).
        The dashed lined box indicates the feature extractor $g_\theta$ in \cref{fig:TDM}.
        For each instance, IAM receives an intermediate feature map $\overline{F}$ and computes the intra score $\overline{R}^\text{intra}$~(distance between channels and salient object regions).
        Then, it produces a channel weight vector $\overline{w}$ that possesses high values in object-relevant channels of the instance.
        By multiplying the intermediate feature map and the weight vector, an adaptive feature map $\overline{A}$ is obtained.
        Instead of forwarding the original feature map, the adaptive feature map is passed to the next layer.
}
Since the TDM is developed for high-level feature maps such as the last convolution layer, its performance can be dependent on the quality of given features produced by the feature extractor. In this section, we further introduce the instance attention module (IAM) by extending the QAM to reflect our motivation even for the intermediate feature representations. Specifically, the IAM is designed to highlight the object-relevant channels for each instance regardless of the support or query sets.

\Skip{
            Although TDM is a well-tailored module for fine-grained few-shot classification and shows superior performance as described in \cref{sec:exp}, it can be affected by the quality of the feature map extracted by the feature extractor.
            In this section, thus, we further extend QAM to produce the feature map with more object-relevant information for each instance.
            Since SAM highly assumes the existence of the support set, it is not proper to handle a unit of instance.
            Similarly, QAM is also devised for the query set, but it computes every weight vector only with a single query instance.
            Therefore, QAM can be extended in an instance-wise manner to enhance the quality of feature representation.
}

\begin{table}[t!]
    \caption{
    The architecture of the fully-connected blocks, $b^{\text{intra}}$ and  $b^{\text{inter}}$ in \cref{in_out_weight_vector}, and $\overline{b}$ in \cref{eq_iam_w_b}.
    The batch size B and the size of input C are different across SAM, QAM, and IAM.
    For the SAM, B is the number of categories comprising an episode, and C is the number of channels of the feature map $F$. On the other hand, for the QAM, C is the same with the SAM, while B is the number of queries. In the IAM, B and C are the numbers of images and channels of intermediate feature map $\overline{F}$, respectively.
    }
    \centering
    \begin{tabular}{|c|c|c|}
        \hline
        \multicolumn{3}{|c|}{\textbf{Fully Connected Block}}\\
        \hline
        Layer & Input Shape & Output Shape \\
        \hline
        Fully Connected Layer & B $\times$ C & B $\times$ 2C\\
        \hline
        Batch Normalization & B $\times$ 2C & B $\times$ 2C\\
        \hline
        ReLU & B $\times$ 2C & B $\times$ 2C\\
        \hline
        Fully Connected Layer & B $\times$ 2C & B $\times$ C\\
        \hline
        1 + Tanh & B $\times$ C & B $\times$ C\\
        \hline
    \end{tabular}
    \label{fully_connectec_block}
\end{table}

The overall architecture of the IAM is illustrated in \cref{fig:IAM}.
IAM operates in intermediate layers of the feature extractor for each instance separately and aims to emphasize the channels encoding object-relevant features. Specifically, IAM receives an intermediate feature map $\overline{F} \in \mathbb{R}^{\overline{C}{\times}\overline{H}{\times}\overline{W}}$ as input, where $\overline{C}$, $\overline{H}$, and $\overline{W}$ denote the number of channels, height, and width of the feature map, respectively.
Then, the channel-wise representativeness score is then defined within the feature map, $\overline{R}^{\text{intra}}_{c}$, for $c$-th channel, as follows:
\begin{equation}
    \label{eq:iam_intra}
    \overline{R}^{\text{intra}}_{c} = \frac{1}{\overline{H} \times \overline{W}} \parallel \overline{f}_{c} - \overline{M} \parallel^2,
\end{equation}
where $\overline{f}_c$ and $\overline{M}$ are the $c$-th channel of $\overline{F}$ and the mean spatial feature computed by the channel-wise average of $\overline{F}$, respectively.
Based on the score vector $\overline{R}^\text{intra} \in \mathbb{R}^{\overline{C}}$, IAM infers a channel weight vector
$\overline{w} \in \mathbb{R}^{\overline{C}} $
in a similar way to QAM described in \cref{query_weight_vector} as follows:
\begin{equation}
\label{eq_iam_w_b}
    \overline{w}=\overline{b} \left(\overline{R}^{\text{intra}}\right),
\end{equation}
where $\overline{b}$ is a fully-connected block as described in \cref{fully_connectec_block}.
Subsequently, an intermediate feature map $\overline{F}$ is then transformed into an attentive feature map $\overline{A}$ based on the channel weight vector $\overline{w}$ as follows:
\begin{equation}
\label{IAMapplication}
    \overline{A} =  \left[{\overline{w}_1}{\overline{f}_1},{\overline{w}_2}{\overline{f}_2}, ... , {\overline{w}_{\overline{C}}}{\overline{f}_{\overline{C}}} \right],
\end{equation}
where $\overline{w}_{i}$ is a scalar at $i$-th dimension of $\overline{w}$, and $\overline{f}_i$ represents $i$-th channel of $\overline{F}$. Finally, the transformed feature map $\overline{A}$ is fed to the next layer in the feature extractor.


In IAM, the computation and application of the channel weight vectors are only performed within each instance; there is no consideration of the support and query sets in the episodic training. Thus, the IAM is applied instance-wisely to all the support and query instances to improve the quality of the feature map. Furthermore, the operations involved in the IAM is conducted only within an instance, the additional computational and memory overhead is small, thereby allowing its usage in the intermediate blocks of the backbone.
\section{Experiments}
\label{sec:exp}
In this section, we evaluate the proposed TDM on fine-grained classification benchmarks, and further verify the generalization capability of IAM on the both fine- and coarse-grained benchmarks.
Throughout the tables in this section, we use $\dagger$ to denote a reproduced version of the baselines.

\subsection{Implementation Details}
\noindent\textbf{Baselines.} 
To verify the effectiveness and adaptability of TDM and IAM in fine-grained classification problem, we apply it to various existing methods including ProtoNet~\cite{protonet}, DSN\cite{dsn}, CTX\cite{ctx}, and FRN\cite{frn}.
On the other hand, for coarse-grained classification problem, we attach IAM to the ProtoNet, FRN, and DeepBDC~\cite{deepbdc}.
For a fair comparison, we reproduce each baseline model with the same hyperparameter described in FRN and DeepBDC.
And, the same training and evaluation scheme is utilized whether TDM or IAM is applied or not.
While TDM generally exploits the prototype\cite{protonet} defined in \cref{prototype} for computing the intra and inter scores, it instead utilizes a query-aligned prototype proposed in the CTX~\cite{ctx} when combining with the CTX.

\noindent\textbf{Architecture.} 
We adopt model architectures commonly utilized in the recent few-shot classification literature \cite{kim2019edge, chen2021pareto, zhao2021looking, hong2021reinforced, zhang2021rethinking}; we employ Conv-4 and ResNet-12.
While both backbone networks accept an image of size 84$\times$84, the size of feature maps is different according to the backbone network. Specifically, 
ResNet-12 yields a feature map with dimensions of 640$\times$5$\times$5, while Conv-4 produces 64$\times$5$\times$5 shape..
For our proposed TDM and IAM, we additionally utilize fully-connected layer blocks where the size of blocks are proportional to the number of channels of the feature maps as described in \cref{fully_connectec_block}. 
We attach IAM to the first and second blocks.
The $\alpha, \beta$ in \cref{support_weight_vector}, \cref{task_weight_vector} are fixed to 0.5.

\noindent\textbf{Training Details.} 
Following the baseline methods~\cite{closer, wang2019simpleshot, feat, deepemd, frn}, we use standard data augmentation techniques including random crop, horizontal flip, and color jitter. 
To prevent overfitting, we add random noise between $-0.2$ and $0.2$ to each output of TDM and IAM. We also regulate the each output of our modules to be in a range of $[0,2]$.
The hyperparameter and training details are followed our baselines for a fair comparison regardless of use of TDM or IAM.

\noindent\textbf{Evaluation Details.} 
For the 5-way $K$-shot experiments, we conduct the evaluation with 10,000 randomly sampled episodes which contain 16 queries per class.
We report average classification accuracy with 95\% confidence intervals.
The 1-shot performances of DSN and CTX are measured by models trained by 5-shot episodes since it shows better performance like FRN.

\begin{table}[t]
    \caption{
	The splits of datasets. While $C_{\text{all}}$ is the number of total classes, $C_{\text{train}}$, $C_{\text{val}}$, $C_{\text{test}}$ are the number of training, validation, and test classes, respectively.
	The classes of these subsets are disjoint.
	}
    \centering{
    \begin{tabular}{l | c c c c}
        \hlineB{2.5}
        \multicolumn{1}{l}{\textbf{Dataset}} & \textbf{$C_{\text{all}}$} & \textbf{$C_{\text{train}}$} & \textbf{$C_{\text{val}}$} & \textbf{$C_{\text{test}}$} \\
        \hlineB{2.5}
        \multicolumn{1}{l}{CUB-200-2011} & 200 & 100 & 50 & 50 \\
        \multicolumn{1}{l}{Aircraft} & 100 & 50 & 25 & 25 \\
        \multicolumn{1}{l}{meta-iNat} & 1135 & 908 & - & 227 \\
        \multicolumn{1}{l}{tiered meta-iNat} & 1135 & 781 & - & 354 \\
        \multicolumn{1}{l}{Stanford-Cars} & 196 & 130 & 17 & 49 \\
        \multicolumn{1}{l}{Stanford-Dogs} & 120 & 70 & 20 & 30 \\
        \multicolumn{1}{l}{Oxford-Pets} & 37 & 20 & 7 & 10 \\
        \multicolumn{1}{l}{mini-ImageNet} & 100 & 64 & 20 & 16 \\
        \multicolumn{1}{l}{tiered-ImageNet} & 608 & 351 & 97 & 160 \\
        \hlineB{2.5}
    \end{tabular}
    }
    \label{tab:dataset_split}
\end{table}
\subsection{Datasets}
We use seven benchmarks for fine-grained few-shot classification: CUB-200-2011, Aircraft, meta-iNat, tiered meta-iNat, Stanford-Cars, Stanford-Dogs, and Oxford-Pets. 
For the evaluation in coarse-grained scenarios, mini-ImageNet and tiered-ImageNet are used.
The split information of each dataset is reported in \cref{tab:dataset_split}.

\noindent\textbf{CUB-200-2011}\cite{wah2011caltech} comprises 11,788 photos of 200 bird species.
This dataset can be utilized in two types: raw form\cite{closer} or preprocessed form by a human-annotated bounding box\cite{feat, deepemd}.
In our work, experiments are conducted with both forms as did in \cite{closer, frn}.

\noindent\textbf {Aircraft}\cite{maji2013fine} is a dataset with 10,000 images of 100 airplane classes.
The main challenge of this dataset arises from airline symbols. Specifically, although the aircraft models are different, their airline symbol can be the same. 
It makes the recognition task more difficult.
Our protocols in splitting the train/test data and image preprocessing with the bounding box are following a way of our baseline model, FRN.

\noindent\textbf{meta-iNat}\cite{wertheimer2019few, van2018inaturalist} 
is a long-tailed dataset.
It contains 1,135 animal species, and the number of images for each category is non-uniform and ranging between 50 and 1000.
For train and test data split, we adopt the way introduced in \cite{wertheimer2019few} which initially proposed this benchmark for the few-shot classification.
However, unlike \cite{wertheimer2019few} where a 227-way evaluation scheme is employed, we adopt a standard 5-way evaluation scheme following our baseline model, FRN.

\noindent\textbf{tiered meta-iNat}\cite{wertheimer2019few} has the same images with meta-iNat.
However, the difference comes from how the train and test data are organized; unlike meta-iNat, the tiered version divides the split by super categories.
Therefore, a bigger domain gap exists between train and test classes.

\noindent\textbf{Stanford Cars}\cite{krause20133d} consists of 16,185 images of 196 car classes.
We employ the same data split protocol with \cite{li2019revisiting} that first introduced this dataset for the few-shot classification task.

\begin{table}[t]
    \caption{
	Performance on CUB using bounding-box cropped images as input. 
	``$\ast$" denotes reproduced one in RENet. 
	Confidence intervals for our implemented model are all below 0.23.
	}
    \centering
    \begin{tabular}{l | c c c c}
        \hlineB{2.5}
        \multicolumn{1}{l}{\multirow{2}{*}{\textbf{Model}}} & \multicolumn{2}{c}{\textbf{Conv-4}} & \multicolumn{2}{c}{\textbf{ResNet-12}} \\
        \multicolumn{1}{c}{}& \textbf{1-shot} & \textbf{5-shot} & \textbf{1-shot} & \textbf{5-shot} \\
        \hlineB{2.5}
        \multicolumn{1}{l}{MatchNet~\cite{matchnet, feat, deepemd}} & 67.73 & 79.00 & 71.87 & 85.08 \\
        \multicolumn{1}{l}{FEAT{$^{\ast}$}~\cite{feat}} & 68.87 & 82.90 & 73.27 & 85.77 \\
        \multicolumn{1}{l}{DeepEMD~\cite{deepemd}} & - & - & 75.65 & 88.69 \\
        \multicolumn{1}{l}{RENet~\cite{kang2021relational}} & - & - & 79.49 & 91.11 \\
        \hlineB{1.0}
        \multicolumn{1}{l}{ProtoNet{$^{\dagger}$}~\cite{protonet}} & 62.90 & 84.13 & 78.99 & 90.74 \\
        \multicolumn{1}{l}{~~~+ TDM} & 69.94 & 86.96 & 79.58 & 91.28 \\
        \multicolumn{1}{l}{~~~+ IAM} & 68.18 & 85.96 & 79.65 & 91.20 \\
        \multicolumn{1}{l}{~~~+ TDM + IAM} & 72.96 & 88.02 & 80.93 & 91.80
        \\
        \hlineB{1.}
        \multicolumn{1}{l}{DSN{$^{\dagger}$}~\cite{dsn}} & 72.09 & 85.03 & 80.51 & 90.23 \\
        \multicolumn{1}{l}{~~~+ TDM} & 73.38 & 86.07 & 81.33 & 90.65 \\
        \multicolumn{1}{l}{~~~+ IAM} & 75.10 & 86.66 & 82.03 & 90.67 \\
        \multicolumn{1}{l}{~~~+ TDM + IAM} & 74.75 & 86.89 & 82.85 & 91.47
        \\
        \hlineB{1.}
        \multicolumn{1}{l}{CTX{$^{\dagger}$}~\cite{ctx}} & 72.14 & 87.23 & 80.67 & 91.55 \\
        \multicolumn{1}{l}{~~~+ TDM} & 74.68 & 88.36 & 83.28 & 92.74 \\
        \multicolumn{1}{l}{~~~+ IAM} & 75.65 & 89.07 & 82.87 & 92.49 \\
        \multicolumn{1}{l}{~~~+ TDM + IAM} & \textbf{77.17} & \textbf{89.90} & 83.76 & 92.85 \\
        \hlineB{1.}
        \multicolumn{1}{l}{FRN{$^{\dagger}$}~\cite{frn}} & 73.24 & 88.33 & 83.16 & 92.42 \\
        \multicolumn{1}{l}{~~~+ TDM} & 74.39 & 88.89 & 83.36 & 92.80 \\
        \multicolumn{1}{l}{~~~+ IAM} & 76.29 & 89.66 & 83.63 & 92.59 \\
        \multicolumn{1}{l}{~~~+ TDM + IAM} & 75.49 & 89.72 & \textbf{84.17} & \textbf{93.30}
        \\
        \hlineB{2.5}
    \end{tabular}
    \label{CUB_cropped}
\end{table}
\noindent\textbf{Stanford Dogs}\cite{khosla2011novel} contains 20,580 images belonging to one of 120 breeds of dogs around the world.
Similar to Stanford Cars, it is also introduced by \cite{li2019revisiting} for fine-grained few-shot classification.
Thus, we follow \cite{li2019revisiting} in the way of splitting this dataset.

\noindent\textbf{Oxford Pets}\cite{parkhi2012cats} is another fine-grained image dataset that has 37 pet classes with approximately 200 images per category.
To the best of our knowledge, this dataset has never been used for the few-shot classification task before the previous conference version of this paper~\cite{tdm}. Thus, we randomly divide classes to define the train/test split as did in \cite{tdm}.

\noindent\textbf{mini-ImageNet}~\cite{matchnet} is one of the representative benchmarks for the few-shot classification.
It is a subset of ImageNet and comprises 100 classes where 600 different images exist per category.
Our dataset split is adopted from \cite{matchnet}.
Unlike the aforementioned datasets which are utilized to validate the effectiveness of TDM and IAM for fine-grained classification, we use mini-ImageNet to evaluate the generalization capability of IAM.

\noindent\textbf{tiered-ImageNet}~\cite{tiered_imagenet} is also a subset of ImageNet, but it is the largest dataset for the the few-shot classification.
It contains 601 categories which is much greater than the number of classes of mini-ImageNet.
Moreover, unlike mini-ImageNet, this benchmark separates train, evaluation, and test classes by super categories, therefore, a large domain gap exists like tiered meta-iNat.
We utilize this benchmark to validate the IAM, since it is a coarse-grained classification dataset.

\begin{table}[t]
    \caption{
    Performance on CUB using raw images as input. 
    }
    \centering
    \begin{tabular}{l | c c c}
        \hlineB{2.5}
        \multicolumn{1}{l}{\textbf{Model}} & \textbf{Backbone} & \textbf{1-shot} & \textbf{5-shot} \\
        \hlineB{2.5}
        \multicolumn{1}{l}{Baseline~\cite{closer}} & ResNet-18 & 65.51$\pm$0.87 & 82.85$\pm$0.55 \\
        \multicolumn{1}{l}{Baseline++~\cite{closer}} & ResNet-18 & 67.02$\pm$0.90 & 83.58$\pm$0.54 \\
        \multicolumn{1}{l}{MatchNet~\cite{closer, matchnet}} & ResNet-18 & 73.42$\pm$0.89 & 84.45$\pm$0.58 \\
        \multicolumn{1}{l}{MAML~\cite{closer, maml}} & ResNet-18 & 68.42$\pm$1.07 & 83.47$\pm$0.62 \\
        \multicolumn{1}{l}{RelatioNet~\cite{closer, sung2018learning}} & ResNet-18 & 68.58$\pm$0.94 & 84.05$\pm$0.56 \\
        \multicolumn{1}{l}{S2M2~\cite{s2m2}} & ResNet-18 & 71.43$\pm$0.28 & 85.55$\pm$0.52 
        \\
        \multicolumn{1}{l}{Neg-Cosine~\cite{neg_cosine}} & ResNet-18 & 72.66$\pm$0.85 & 89.40$\pm$0.43 
        \\
        \multicolumn{1}{l}{Afrasiyabi \textit{et al.}~\cite{afrasiyabi2020associative}} & ResNet-18 & 74.22$\pm$1.09 & 88.65$\pm$0.55 
        \\
        \hlineB{1.}
        \multicolumn{1}{l}{ProtoNet{$^{\dagger}$}~\cite{protonet}} & ResNet-12 & 78.58$\pm$0.22 & 89.83$\pm$0.12 
        \\
        \multicolumn{1}{l}{~~~+ TDM} & ResNet-12 & 79.11$\pm$0.22 & 90.83$\pm$0.11
        \\
        \multicolumn{1}{l}{~~~+ IAM} & ResNet-12 & 78.28$\pm$0.22 & 90.72$\pm$0.12
        \\
        \multicolumn{1}{l}{~~~+ TDM + IAM} & ResNet-12 & 78.89$\pm$0.22 & 90.86$\pm$0.12
        \\
        \hlineB{1.}
        \multicolumn{1}{l}{DSN{$^{\dagger}$}~\cite{dsn}} & ResNet-12 & 80.47$\pm$0.20 & 89.92$\pm$0.12 
        \\
        \multicolumn{1}{l}{~~~+ TDM} & ResNet-12 & 80.58$\pm$0.20 & 89.95$\pm$0.12
        \\
        \multicolumn{1}{l}{~~~+ IAM} & ResNet-12 & 81.33$\pm$0.20 & 89.87$\pm$0.12
        \\
        \multicolumn{1}{l}{~~~+ TDM + IAM} & ResNet-12 & 81.96$\pm$0.20 & 90.54$\pm$0.12
        \\
        \hlineB{1.}
        \multicolumn{1}{l}{CTX{$^{\dagger}$}~\cite{ctx}} & ResNet-12 & 80.95$\pm$0.21 & 91.54$\pm$0.11
        \\
        \multicolumn{1}{l}{~~~+ TDM} & ResNet-12 & 83.45$\pm$0.19 & 92.49$\pm$0.11
        \\
        \multicolumn{1}{l}{~~~+ IAM} & ResNet-12 & 81.97$\pm$0.20 & 92.04$\pm$0.11
        \\
        \multicolumn{1}{l}{~~~+ TDM + IAM} & ResNet-12 & 83.82$\pm$0.19 & 92.79$\pm$0.10
        \\
        \hlineB{1.}
        \multicolumn{1}{l}{FRN{$^{\dagger}$}\cite{frn}} & ResNet-12 & 83.54$\pm$0.19 & 92.96$\pm$0.10
        \\
        \multicolumn{1}{l}{~~~+ TDM} & ResNet-12 & 84.36$\pm$0.19 & 93.37$\pm$0.10
        \\
        \multicolumn{1}{l}{~~~+ IAM} & ResNet-12 & 84.50$\pm$0.19 & 93.21$\pm$0.10
        \\
        \multicolumn{1}{l}{~~~+ TDM + IAM} & ResNet-12 & \textbf{84.84$\pm$0.18} & \textbf{93.60$\pm$0.10}
        \\
        \hlineB{2.5}
    \end{tabular}
    \label{CUB_raw}
\end{table}

\subsection{Fine-grained Few-Shot Classification}
\begin{table}[t]
    \centering
    \caption{Performance on Aircraft. Confidence intervals for our implemented model are all below 0.25.}
    \begin{tabular}{l | c c c c}
        \hlineB{2.5}
        \multicolumn{1}{l}{\multirow{2}{*}{\textbf{Model}}} & \multicolumn{2}{c}{\textbf{Conv-4}} & \multicolumn{2}{c}{\textbf{ResNet-12}} \\
        \multicolumn{1}{c}{} & \textbf{1-shot} & \textbf{5-shot} & \textbf{1-shot} & \textbf{5-shot} \\
        \hlineB{2.5}
        \multicolumn{1}{l}{ProtoNet{$^{\dagger}$}\cite{protonet}} & 47.37 & 68.96 & 67.28 & 83.21 \\
        \multicolumn{1}{l}{~~~+ TDM} & 50.55 & 71.12 & 69.12 & 84.77 
        \\
        \multicolumn{1}{l}{~~~+ IAM} & 49.67 & 68.57 & 69.10 & 84.04
        \\
        \multicolumn{1}{l}{~~~+ TDM + IAM} & 52.88 & 72.81 & 69.80 & \textbf{85.41}
        \\
        \hlineB{1.}
        \multicolumn{1}{l}{DSN{$^{\dagger}$}\cite{dsn}} & 52.22 & 68.75 & 70.23 & 83.05 
        \\
        \multicolumn{1}{l}{~~~+ TDM} & 53.77 & 69.56 & 71.57 & 83.65 
        \\
        \multicolumn{1}{l}{~~~+ IAM} & 54.62 & 68.87 & 72.01 & 83.36
        \\
        \multicolumn{1}{l}{~~~+ TDM + IAM} & 54.64 & 70.34 & \textbf{73.83} & 85.11
        \\
        \hlineB{1.}
        \multicolumn{1}{l}{CTX{$^{\dagger}$}\cite{ctx}} & 51.58 & 68.12 & 65.53 & 79.31 
        \\
        \multicolumn{1}{l}{~~~+ TDM} & 55.15 & 70.45 & 69.42 & 83.25 
        \\
        \multicolumn{1}{l}{~~~+ IAM} & 54.70 & 70.61 & 70.93 & 82.38
        \\
        \multicolumn{1}{l}{~~~+ TDM + IAM} & \textbf{57.04} & 72.46 & 71.40 & 84.12
        \\
        \hlineB{1.}
        \multicolumn{1}{l}{FRN{$^{\dagger}$}\cite{frn}} & 53.12 & 70.84 & 69.58 & 82.98 
        \\
        \multicolumn{1}{l}{~~~+ TDM} & 54.21 & 71.37 & 70.89 & 84.54 
        \\
        \multicolumn{1}{l}{~~~+ IAM} & 54.98 & 72.12 & 71.23 & 83.66
        \\
        \multicolumn{1}{l}{~~~+ TDM + IAM} & 56.08 & \textbf{72.62} & 72.36 & 85.05
        \\
        \hlineB{2.5}
    \end{tabular}
    \label{aircraft}
\end{table}
\noindent\textbf{CUB-200-2011 results.} 
\begin{figure*}[h]
    \centering
    \includegraphics[width=.95\textwidth]{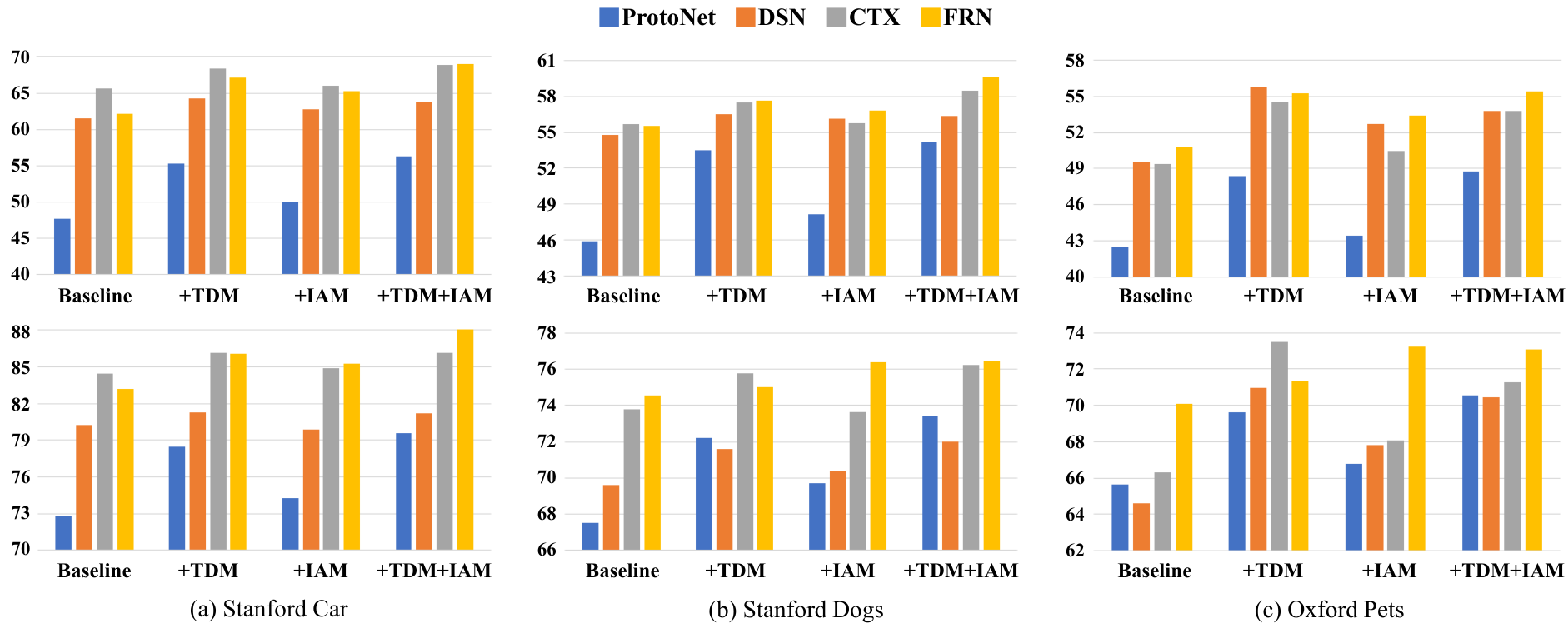}
    \caption{
    Experimental results on (a) Stanford Car, (b) Stanford Dogs, and (c) Oxford Pets. 
    The above and below graphs for each dataset indicate 1-shot and 5-shot performances, respectively.
    The bars at the first column of each graph report the accuracies of baselines.
    And, the bars in the second and third columns indicate the accuracies when TDM and IAM are combined with baselines, respectively.
    The bars in the last column represent the performances where TDM and IAM are utilized together.
    }
    \label{fig:additional_dataset}
\end{figure*}
\cref{CUB_cropped} and \cref{CUB_raw} report the results of our baselines and their performances when our proposed modules, TDM and IAM, are combined.
With cropped images in \cref{CUB_cropped}, TDM and IAM consistently improve the performance of baseline models in all cases and achieve the state-of-the-art scores when they are applied together.
Despite a few settings when IAM is not much effective in \cref{CUB_raw}, TDM and IAM together, still show superior performances.

\noindent\textbf {Aircraft results.} 
As reported in \cref{aircraft}, TDM and IAM show a consistent tendency by boosting performances of all the baselines regardless of the model size and the size of support sets, except for one case.
Although there is a slight performance decrease when IAM is used in the 5-shot scenario of ProtoNet with Conv-4 backbone, more outstanding performance is achieved when IAM and TDM are adopted together, compared to utilizing TDM only.

\noindent\textbf{meta-iNat and tiered meta-iNat results.} 
These datasets are suitable for evaluation of model's generalization capability, since it is widely known that models trained on those datasets are vulnerable to overfitting due to the absence of validation set~\cite{james2013introduction, montavon2012neural, xu2018splitting}.
Moreover, the tiered meta-iNat makes the task more difficult since super categories of train and test set do not overlap.
As reported in \cref{inat}, TDM and IAM are robust to the overfitting and the large domain gap as they enhance the results in most configurations.
For a slight performance decrease of TDM combined with FRN in a 5-shot scenario on tiered meta-iNat, we believe that this is mainly due to the learnable parameter $\lambda$ in FRN.
In general, large $\lambda$ shows good performances when a domain gap exists.
Yet, we found that TDM restrains the $\lambda$ to be relatively small since TDM assists to focus on discriminative channels.
Therefore, except for the above case, we accomplish the state-of-the-art performances when TDM and IAM are utilized together.

\begin{table}[t]
    \centering
    \caption{
	Performance on meta-iNat and tiered meta-iNat with Conv-4 backbones.
	Confidence intervals for our implemented model are all below 0.23.
    }
    \begin{tabular}{l | c c c c}
        \hlineB{2.5}
        \multicolumn{1}{l}{\multirow{2}{*}{\textbf{Model}}} & \multicolumn{2}{c}{\textbf{meta-iNat}} & \multicolumn{2}{c}{\textbf{tiered meta-iNat}} \\
        \multicolumn{1}{l}{} & \textbf{1-shot} & \textbf{5-shot} & \textbf{1-shot} & \textbf{5-shot} \\
        \hlineB{2.5}
        \multicolumn{1}{l}{ProtoNet{$^{\dagger}$}\cite{protonet}} & 55.37 & 76.30 & 34.41 & 57.60 
        \\
        \multicolumn{1}{l}{~~~+ TDM} & 61.82 & 79.95 & 38.30 & 61.18 
        \\
        \multicolumn{1}{l}{~~~+ IAM} & 59.12 & 79.83 & 37.94 & 63.47
        \\
        \multicolumn{1}{l}{~~~+ TDM + IAM} &65.10 & 81.93 & 41.87 & 64.32
        \\
        \hlineB{1.}
        \multicolumn{1}{l}{DSN{$^{\dagger}$}\cite{dsn}} & 60.06 & 76.15 & 40.83 & 58.34 
        \\
        \multicolumn{1}{l}{~~~+ TDM} & 61.87 & 78.07 & 41.00 & 58.66 
        \\
        \multicolumn{1}{l}{~~~+ IAM} & 63.41 & 77.76 & 44.05 & 61.45
        \\
        \multicolumn{1}{l}{~~~+ TDM + IAM} & 62.99 & 78.84 & 43.39 & 61.69
        \\
        \hlineB{1.}
        \multicolumn{1}{l}{CTX{$^{\dagger}$}\cite{ctx}} & 60.80 & 78.57 & 42.24 & 60.54
        \\
        \multicolumn{1}{l}{~~~+ TDM} & 63.26 & 80.75 & 43.90 & 62.29 
        \\
        \multicolumn{1}{l}{~~~+ IAM} & 63.80 & 80.97 & 45.87 & 64.92
        \\
        \multicolumn{1}{l}{~~~+ TDM + IAM} & 64.96 & 81.89 & \textbf{47.40} & 66.12
        \\
        \hlineB{1.}
        \multicolumn{1}{l}{FRN{$^{\dagger}$}\cite{frn}} & 61.98 & 80.04 & 43.95 & 63.45 
        \\
        \multicolumn{1}{l}{~~~+ TDM} & 63.97 & 81.60 & 44.05 & 62.91 
        \\
        \multicolumn{1}{l}{~~~+ IAM} & 65.11 & 82.43 & 47.33 & \textbf{67.48}
        \\
        \multicolumn{1}{l}{~~~+ TDM + IAM} & \textbf{65.95} & \textbf{83.30} & 46.45 & 66.55
        \\
        \hlineB{2.5}
    \end{tabular}
    \label{inat}
\end{table}

\begingroup
\setlength{\tabcolsep}{14pt} 
\renewcommand{\arraystretch}{1.0} 
\begin{table*}[t]
    \centering
    \caption{
    Performances of IAM on mini-ImageNet and tiered-ImageNet.
    FRN-EMD denotes the FRN implemented in the DeepEMD.
    }
    \label{mini_imagenet}
    \begin{tabular}{l | c c c c c c}
        \hlineB{2.5}
        \multicolumn{1}{l}{\multirow{2}{*}{\textbf{Model}}} & 
        \multicolumn{1}{l}{\multirow{2}{*}{\textbf{Backbone}}} & 
        \multicolumn{2}{c}{\textbf{mini-ImageNet}} &
        \multicolumn{2}{c}{\textbf{tiered-ImageNet}}
        \\
        \multicolumn{1}{l}{} &
        \multicolumn{1}{l}{} &
        \multicolumn{1}{c}{\textbf{1-shot}} &
        \multicolumn{1}{c}{\textbf{5-shot}} &
        \multicolumn{1}{c}{\textbf{1-shot}} &
        \multicolumn{1}{c}{\textbf{5-shot}}
        \\
        \hlineB{2.5}
        \multicolumn{1}{l}{MatchNet~\cite{matchnet, feat, deepemd}} & ResNet-12 & 64.64$\pm$0.20 & 78.72$\pm$0.15 & 68.50$\pm$0.92 & 80.60$\pm$0.71
        \\
        \multicolumn{1}{l}{Baseline++~\cite{closer}} & ResNet-12 & 60.56$\pm$0.45 & 77.40$\pm$0.34 & - & -
        \\
        \multicolumn{1}{l}{CTM~\cite{ctm}} & ResNet-18 & 64.14$\pm$0.82 & 80.51$\pm$0.13 & 68.41$\pm$0.39 & 84.28$\pm$1.73
        \\
        \multicolumn{1}{l}{TADAM~\cite{tadam}} & ResNet-12 & 58.50$\pm$0.30 & 76.70$\pm$0.38 & - & -
        \\
        \multicolumn{1}{l}{S2M2~\cite{s2m2}} & ResNet-12 & 64.06$\pm$0.18 & 80.58$\pm$0.12 & - & -
        \\
        \multicolumn{1}{l}{Neg-Cosine~\cite{neg_cosine}} & ResNet-12 & 63.85$\pm$0.81 & 81.57$\pm$0.56 & - & -    
        \\
        \multicolumn{1}{l}{Afrasiyabi \textit{et al.}~\cite{afrasiyabi2020associative}} & ResNet-12 & 59.88$\pm$0.67 & 80.35$\pm$0.73 & 69.29$\pm$0.56 & 85.97$\pm$0.49
        \\
        \multicolumn{1}{l}{FEAT~\cite{feat}} & ResNet-12 & 66.78$\pm$0.20 & 82.05$\pm$0.14 & 70.80$\pm$0.23 & 84.79$\pm$0.16
        \\
        \multicolumn{1}{l}{DeepEMD~\cite{deepemd}} & ResNet-12 & 65.91$\pm$0.82 & 82.41$\pm$0.56 & 71.16$\pm$0.87 & 86.03$\pm$0.58
        \\
        \hlineB{1.}
        \multicolumn{1}{l}{ProtoNet$^\dagger$~\cite{protonet, deepbdc}} & ResNet-12 & 61.72$\pm$0.20 & 78.75$\pm$0.14 & - & -
        \\
        \multicolumn{1}{l}{~+ IAM} & ResNet-12 & 62.25$\pm$0.20 & 79.44$\pm$0.15 & - & -
        \\
        \hlineB{1.}
        \multicolumn{1}{l}{FRN$^\dagger$~\cite{frn}} & ResNet-12 & 66.69$\pm$0.19 & 82.89$\pm$0.13 & 71.13$\pm$0.22 & 86.13$\pm$0.15
        \\
        \multicolumn{1}{l}{~+ IAM} & ResNet-12 & 66.96$\pm$0.19 & 83.19$\pm$0.13 & 71.85$\pm$0.22 & 86.55$\pm$0.15
        \\
        \multicolumn{1}{l}{FRN-EMD$^{{\dagger}}$~\cite{frn, deepemd}} & ResNet-12 & - & - & 72.15$\pm$0.22 & 86.49$\pm$0.15
        \\
        \multicolumn{1}{l}{~+ IAM} & ResNet-12 & - & - & \textbf{72.84$\pm$0.22} & \textbf{87.04$\pm$0.14}
        \\
        \hlineB{1.}
        \multicolumn{1}{l}{Meta DeepBDC$^\dagger$~\cite{deepbdc}} & ResNet-12 & 65.74$\pm$0.20 & 83.23$\pm$0.13 & - & -
        \\
        \multicolumn{1}{l}{~+ IAM} & ResNet-12 & 66.21$\pm$0.20 & 83.78$\pm$0.13 & - & -
        \\
        \multicolumn{1}{l}{STL DeepBDC$^\dagger$~\cite{deepbdc}} & ResNet-12 & 67.62$\pm$0.20 & 84.65$\pm$0.13 & -	& -
        \\
        \multicolumn{1}{l}{~+ IAM} & ResNet-12 & \textbf{67.95$\pm$0.19} & \textbf{84.86$\pm$0.13} & - & -
        \\
        \hlineB{2.5}
    \end{tabular}
\end{table*}
\endgroup

\begin{table}[t]
    \caption{
    Cross-domain few-shot classification performance of a scenario where models are trained with mini-ImageNet and tested on CUB.
    }
    \centering
    \begin{tabular}{l | c c c}
        \hlineB{2.5}
        \multicolumn{1}{l}{\textbf{Model}} & \textbf{Backbone} & \textbf{1-shot} & \textbf{5-shot} \\
        \hlineB{2.5}
        \multicolumn{1}{l}{Baseline~\cite{closer}} & ResNet-18 & - & 51.34$\pm$0.72 \\
        \multicolumn{1}{l}{Baseline++~\cite{closer}} & ResNet-18 & - & 62.02$\pm$0.70 \\
        \multicolumn{1}{l}{MAML~\cite{closer, maml}} & ResNet-18 & - & 51.34$\pm$0.72 \\
        \multicolumn{1}{l}{Afrasiyabi \textit{et al.}~\cite{afrasiyabi2020associative}} & ResNet-18 & 46.85$\pm$0.75 & 70.37$\pm$1.02 
        \\
        \hlineB{1.}
        \multicolumn{1}{l}{ProtoNet{$^{\dagger}$}~\cite{protonet, deepbdc}} & ResNet-12 & 46.58$\pm$0.19 & 66.19$\pm$0.17 
        \\
        \multicolumn{1}{l}{~~~+ IAM} & ResNet-12 & 47.67$\pm$0.19 & 68.70$\pm$0.17
        \\
        \hlineB{1.}
        \multicolumn{1}{l}{FRN{$^{\dagger}$}\cite{frn}} & ResNet-12 & 52.80$\pm$0.21 & 73.75$\pm$0.18
        \\
        \multicolumn{1}{l}{~~~+ IAM} & ResNet-12 & 54.94$\pm$0.22 & 75.76$\pm$0.18
        \\
        \hlineB{1.}
        \multicolumn{1}{l}{Meta DeepBDC{$^{\dagger}$}\cite{deepbdc}} & ResNet-12 & 42.83$\pm$0.20 & 74.11$\pm$0.16
        \\
        \multicolumn{1}{l}{~~~+ IAM} & ResNet-12 & 45.80$\pm$0.21 & \textbf{77.71$\pm$0.15}
        \\
        \multicolumn{1}{l}{STL DeepBDC{$^{\dagger}$}\cite{deepbdc}} & ResNet-12 & 55.01$\pm$0.21 & 75.47$\pm$0.16
        \\
        \multicolumn{1}{l}{~~~+ IAM} & ResNet-12 & \textbf{55.92$\pm$0.20} & 76.43$\pm$0.16
        \\
        \hlineB{2.5}
    \end{tabular}
    \label{cross_domain_cub}
\end{table}

\noindent\textbf{Stanford Cars, Stanford Dogs, and Oxford Pets results.} 
Although these datasets have fine-grained classes, they were not utilized for evaluation of our baselines in the literature.
To further validate the effectiveness of TDM and IAM, we additionally conduct experiments on those datasets with Conv-4.
As shown in \cref{fig:additional_dataset}, TDM and IAM generally improve the performances except for a few cases when IAM is solely used.
In detail, TDM improves the accuracy scores by 4.44 and 3.27\%p compared to the baselines at 1-shot and 5-shot scenarios, respectively.
On the other hand, IAM provides 1.69 and 1.46\%p performance improvements, respectively.
Furthermore, as our modules are compatible with one another, we observe significant boosts when they are both adopted; there are 4.79 and 3.81\%p boosts in accuracy scores in 1-shot and 5-shot cases, respectively.

Throughout the extensive experiments on seven benchmark datasets, we validated the merits of TDM and IAM in the fine-grained classification task.
To summarize the experimental results, TDM has shown its superiority regardless of the datasets and baseline methods. For IAM, although it encourages to highlight object-relevant channels within the feature map, sometimes it is not beneficial for the fine-grained classification since the objects in the fine-grained datasets could have excessive common features across classes.
However, when IAM and TDM are utilized together, this is no longer a problem since TDM highlights class-discriminative features among object-relevant ones identified by the IAM.
On the other side, IAM helps TDM to discover more discriminative features, since IAM provides more object-focused feature maps to the TDM. Therefore, we claim that TDM and IAM have complementary benefits.
\begin{table}[t]
    \caption{
    Cross-domain few-shot classification performance of a scenario where models are trained with mini-ImageNet and tested on Aircraft. Unlike the results on \cref{cross_domain_cub}, STL DeepBDC in this table does not perform distillation stages since skipping these phases shows better performances.
    }
    \centering
    \begin{tabular}{l | c c c}
        \hlineB{2.5}
        \multicolumn{1}{l}{\textbf{Model}} & \textbf{Backbone} & \textbf{1-shot} & \textbf{5-shot} \\
        \hlineB{2.5}
        \multicolumn{1}{l}{ProtoNet{$^{\dagger}$}~\cite{protonet, deepbdc}} & ResNet-12 & 33.48$\pm$0.15 & 49.55$\pm$0.18 
        \\
        \multicolumn{1}{l}{~+ IAM} & ResNet-12 & 34.65$\pm$0.15 & 51.00$\pm$0.18
        \\
        \hlineB{1.}
        \multicolumn{1}{l}{FRN{$^{\dagger}$}\cite{frn}} & ResNet-12 & 38.71$\pm$0.16 & 62.10$\pm$0.18
        \\
        \multicolumn{1}{l}{~+ IAM} & ResNet-12 & \textbf{39.77$\pm$0.17} & \textbf{63.61$\pm$0.18}
        \\
        \hlineB{1.}
        \multicolumn{1}{l}{Meta DeepBDC{$^{\dagger}$}\cite{deepbdc}} & ResNet-12 & 36.11$\pm$0.16 & 59.52$\pm$0.19
        \\
        \multicolumn{1}{l}{~+ IAM} & ResNet-12 & 37.46$\pm$0.17 & 60.66$\pm$0.19
        \\
        \multicolumn{1}{l}{STL DeepBDC{$^{\dagger}$}\cite{deepbdc}} & ResNet-12 & 38.18$\pm$0.17 & 57.61$\pm$0.19
        \\
        \multicolumn{1}{l}{~+ IAM} & ResNet-12 & 38.82$\pm$0.17 & 58.64$\pm$0.19
        \\
        \hlineB{2.5}
    \end{tabular}
    \label{cross_domain_aircraft}
\end{table}

\begin{figure*}[ht]
    \centering
    \includegraphics[width=.95\textwidth]{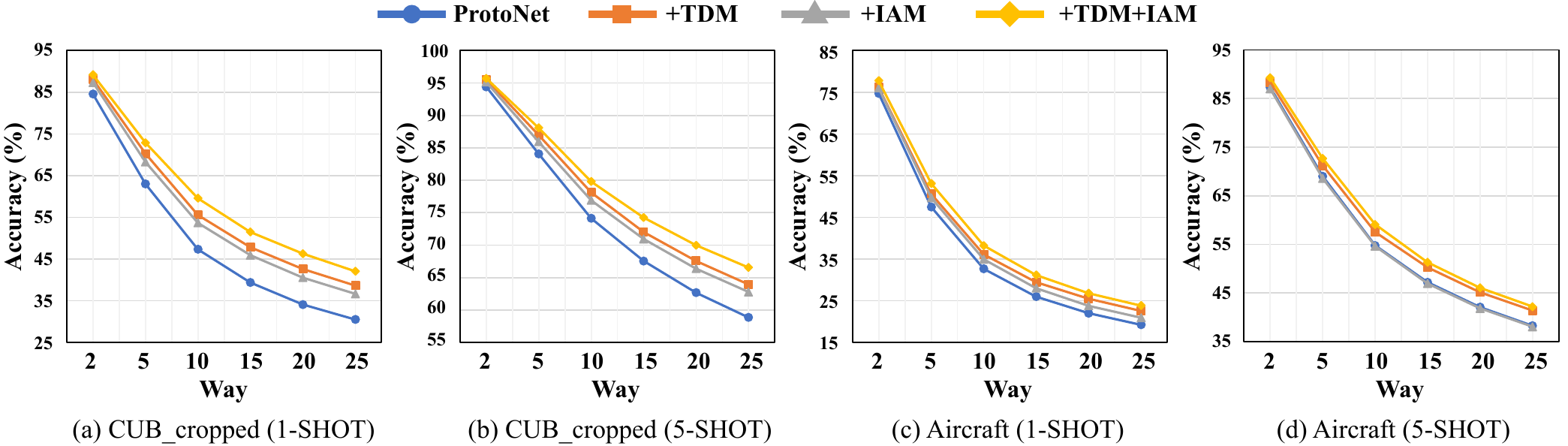}
    \caption{
    Experimental results of N-way 1- and 5-shot classification with varying N.
    }
    \label{fig:n_way}
\end{figure*}
\begin{figure}[t!]
    \centering
    \includegraphics[width=\columnwidth]{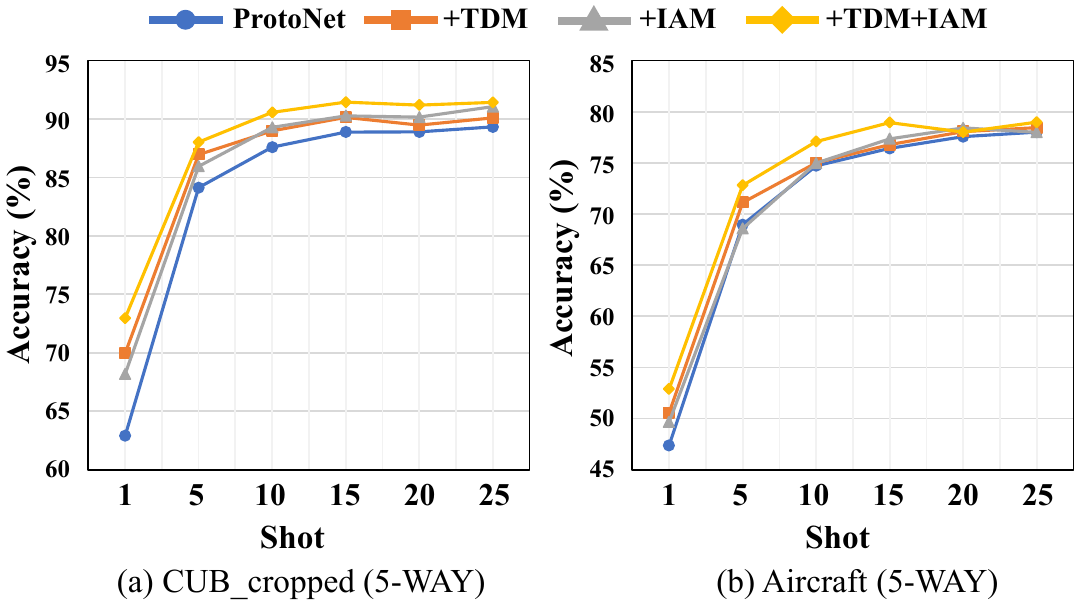}
    \caption{
    5-way K-shot classification results with varying K.
    }
    \label{fig:k_shot}
\end{figure}
\subsection{Coarse-grained Few-Shot Classification}
\noindent\textbf{mini-ImageNet and tiered-ImageNet results.} 
As discussed in \cref{discussion}, TDM is could be not a proper module for the coarse-grained few-shot classification task since it restrains utilizing various features of the object. On the other hand, since IAM encourages the feature extractor to produce various object-relevant features for each instance, we think that IAM is also beneficial for the coarse-grained few-shot classification.
To validate the effectiveness of IAM in coarse-grained benchmarks, we perform experiments on mini-ImageNet and tiered-ImageNet, and the results are reported in \cref{mini_imagenet}.
As can be seen, IAM improves all the baselines regardless of the training scheme.
Besides its effectiveness, we also emphasize the high applicability of IAM, because these results are obtained without any extensive hyperparameter searching or optimizing processes.

\subsection{Cross-domain Few-Shot Classification}
\noindent\textbf{mini-ImageNet $\rightarrow$ CUB-200-2011 results.} 
To evaluate the cross-domain generalization ability of the few-shot classification algorithms, we validate each model when its train and test datasets are different, following the protocol of \cite{closer, frn}.
Since images of fine-grained category are typically collected by professionals in each field, we argue that this setting is deeply related to reducing the cost of labeling.
Specifically, we train each model with mini-ImageNet and validate them with CUB~(raw form), as did in \cite{frn, closer}.
As reported in \cref{cross_domain_cub}, IAM consistently improves the performances of all the baselines and achieves the state-of-the-art without any adaptation process.

\noindent\textbf{mini-ImageNet $\rightarrow$ Aircraft results.}
Although there is a big domain gap between mini-ImageNet and CUB, the train categories of mini-ImageNet still include two bird species as different classes.
Therefore, each model trained with mini-ImageNet, could be already learned to distinguish bird species.
On the other hand, since there are no airplane images in the train set of mini-ImageNet, classifying airplane types is a more proper setting for evaluating the cross-domain generalization capability of models. 
Specifically, we evaluate each model trained with mini-ImageNet on the test set of Airplane dataset. As reported in \cref{cross_domain_aircraft}, our IAM shows its effectiveness even in categories that have never been seen in training stages.
\section{Ablation Study}\label{sec:abl}
In this section, we conduct ablation experiments.
Most experiments for the ablation studies are performed based on  ProtoNet\cite{protonet} with the Conv-4 backbone using CUB\_cropped and Aircraft datasets.

\subsection{Varying N and K for N-way K-shot}
In \cref{sec:exp}, we extensively validate the merits of our proposed modules in various scenarios. However, the number of categories of each episode for those experiments was fixed to 5 by following the protocol of existing work~\cite{matchnet, protonet, frn, deepemd, deepbdc}.
In a real-world scenario, the number of classes can be varying depending on the circumstance. Therefore, to verify the effectiveness of TDM and IAM in the such scenario, we first evaluate our modules with varying number of classes $N$ comprising an episode.
As reported in \cref{fig:n_way}, TDM and IAM provide consistent improvements compared to the baseline, except for one case of 5-shot in Aircraft benchmark.
Moreover, the relative performance improvements are in proportion to the number of categories. It clearly demonstrates that our modules are more effective in more difficult settings (i.e., more classes).

On the other hand, the number of labeled images $K$ of each category was in a range of [1,5] following the existing methods~\cite{closer, feat, kang2021relational, maml}.
Similar to the experiments with respect to the number of classes, we perform experiments with varying numbers of labeled images and the results are provided in  \cref{fig:k_shot}.
As reported, the benefits of our modules are especially highlighted at low-shots in terms of relative performance improvements. It validates that our modules are more suitable for the few-shot scenarios, the main task of this study, while showing their effectiveness in many-shot.

\begingroup
\begin{table}[t]
    \centering
    \caption{Ablation study on SAM, QAM, and IAM.}
    \begin{tabular}{c c c c c c c}
        \hlineB{2.5}
        \multicolumn{1}{c}{\multirow{2}{*}{\textbf{SAM}}} & 
        \multicolumn{1}{c}{\multirow{2}{*}{\textbf{QAM}}} &
        \multicolumn{1}{c}{\multirow{2}{*}{\textbf{IAM}}} & 
        \multicolumn{2}{c}{\textbf{CUB\_cropped}} & 
        \multicolumn{2}{c}{\textbf{Aircraft}} 
        \\
        & & & \textbf{1-shot} & \textbf{5-shot} & \textbf{1-shot} & \textbf{5-shot} 
        \\
        \hlineB{2.5}
        - & - & - & 62.90 & 84.13 & 47.37 & 68.96 
        \\
        \cmark & - & - & 68.53 & 85.95 & 49.45 & 69.33 
        \\
        - & \cmark & - & 65.11 & 84.82 & 48.96 & 70.85 
        \\
        - & - & \cmark & 68.18 & 85.96 & 49.67 & 68.57 
        \\
        \cmark & \cmark & - & 69.94 & 86.96 & 50.55 & 71.12 
        \\
        \cmark & - & \cmark & 71.97 & 88.00 & 51.98 & 70.67
        \\
        - & \cmark & \cmark & 68.87 & 86.68 & 51.69 & 69.60
        \\
        \cmark & \cmark & \cmark & \textbf{72.96} & \textbf{88.02} & \textbf{52.88} & \textbf{72.81}
        \\
        \hlineB{2.5}
    \end{tabular}
    \label{samqam}
\end{table}
\endgroup
\subsection{Ablation Study on SAM, QAM, and IAM}
Since our method consists of three sub-modules, SAM, QAM, and IAM, we perform experiments with various combinations of these sub-modules to evaluate the contribution of each component and confirm their complementary benefits. As reported in \cref{samqam}, each sub-module consistently improves the classification accuracies across the datasets except for one case of IAM. 
The large gains by SAM confirm that identifying and focusing on discriminative channels for each category are crucial for fine-grained few-shot classification. Furthermore, although the improvements by QAM is slightly lower compared to SAM, QAM is also shown to be effective for all tested configurations. It confirms the benefits of applying more importance to the support set features that are possessed by the query instance. On the other hand, the effect of IAM varies from time to time which may degrade the performance as in the 5-shot scenario on Aircraft.
This is because the object-relevant feature maps induced by IAM may hinder accurate predictions when classes share many characteristics. However, as can be observed in the sixth row, SAM is able to resolve the limitation of IAM by restraining common features and discovering discriminative features.
Most importantly, the the best performances are achieved when all three components are utilized together. These results validate the merit of each sub-module and their complementary benefits.

\subsection{Compatibility with the Cosine Distance}
In this paper, we mostly evaluate our method by employing the Euclidean distance when computing the similarity among instances as did in our baselines~\cite{protonet, dsn, ctx, frn}. Meanwhile, the cosine distance is another popular metric adopted in other techniques~\cite{matchnet, closer, metabaseline}. Therefore, we also validate the compatibility of our method with the cosine distance, and results are reported in \cref{cosine_simialrity}. The consistent tendency that adopting IAM or TDM leads to significant performance gains confirms the compatibility of our method with the cosine distance metric.

\subsection{Comparison to Existing Attention Methods}
To further verify the benefits of our method over existing attention modules in the fine-grained few-shot classification task, we compare our method with SENet~\cite{senet}, CBAM~\cite{cbam}, and Self-attention~\cite{selfattention}. As reported in \cref{tab:general_attention_methods}, TDM+IAM outperforms the existing attention methods by large margins. Note that, the main difference between our method and the existing modules is that we explicitly measure and leverage the channel-wise importance based on their representativeness scores in our attention modules, while existing methods are relied on the learnable parameters less suitable for the few-shot and fine-grained scenarios. Consequently, these results confirm that our modules has clear benefits over the existing attention modules in the fine-grained few-shot classification task.
\begingroup
\setlength{\tabcolsep}{9pt} 
\begin{table}[t]
    \centering
    \caption{
    Compatibility with the cosine distance.
    }
    \begin{tabular}{c c c c c c}
        \hlineB{2.5}
        \multicolumn{1}{c}{\multirow{2}{*}{\textbf{Method}}} 
        & \multicolumn{2}{c}{\textbf{CUB\_cropped}} & \multicolumn{2}{c}{\textbf{Aircraft}} 
        \\
        \multicolumn{1}{c}{} & \textbf{1-shot} & \textbf{5-shot} & \textbf{1-shot} & \textbf{5-shot} 
        \\
        \hlineB{2.5}
        \multicolumn{1}{l}{ProtoNet$^{\dagger}$~\cite{protonet}} & 68.69 & 82.89 & 48.36 & 63.45 
        \\
        \hline
        \multicolumn{1}{l}{~+ TDM} & 69.90 & 84.95 & 51.51 & 68.35 \\
        \multicolumn{1}{l}{~+ IAM} & 69.72 & 84.28 & 50.69 & 66.13 \\
        \multicolumn{1}{l}{~+ TDM + IAM} & 71.17 & 85.15 & 52.62 & 69.62 \\
        \hlineB{2.5}
    \end{tabular}
    \label{cosine_simialrity}
\end{table}
\endgroup
\section{Conclusion}\label{sec:con}
In this paper, we first introduced channel attention modules tailored for the fine-grained few-shot image classification, Task Discrepancy Maximization (TDM) with two submodules, Support Attention Module (SAM) and Query Attention Module (QAM). The core principle of the SAM is to emphasize feature map channels encoding class-discriminative information, while one of the QAM is to concentrate object-relevant channels for the query image. These channel attention modules enable to produce task-adaptive feature maps more focusing on the discriminative details to distinguish among fine-grained categories. To further improve the representation capability for both fine- and coarse-grained few-shot classification, we extended the QAM to present the Instance Attention Module (IAM). Specifically, the IAM operates in the intermediate layers to highlight object-relevant channels for each instance regardless of support or query image unlike the QAM which works for high-level feature maps of the query instance. We extensively evaluated the proposed modules on several fine- and coarse-grained image classification benchmarks to validate their unique merits in terms of effectiveness and applicability to the prior few-shot classification methods.
\Skip{
        In this paper, we first introduced Task Discrepancy Maximization (TDM), a tailored module for fine-grained few-shot classification.
        To distinguish similar classes, TDM computes channel weight vectors that emphasize features of fine discriminative detail with two submodules: Support Attention Module (SAM) and Query Attention Module (QAM).
        Then, to further improve the representation quality for both fine- and coarse-grained few shot task, we extended QAM to present IAM.
        Unlike QAM, IAM operates in the intermediate layers by producing and highlighting the object features with an instance-wise channel weight vector.
        Our extensive experiments on several fine- and coarse-grained benchmarks validated the merits of our proposed TDM and IAM in terms of their effectiveness and high applicability with the prior few-shot classification methods.
}
\begingroup
\setlength{\tabcolsep}{10pt} 
\begin{table}[t]
    \centering
    \caption{
    Comparison with existing attention methods. 
    }
    \begin{tabular}{l c c c c}
        \hlineB{2.5}
        \multicolumn{1}{c}{\multirow{2}{*}{\textbf{Method}}} & \multicolumn{2}{c}{\textbf{CUB\_cropped}} & \multicolumn{2}{c}{\textbf{Aircraft}}
        \\
        & \textbf{1-shot} & \textbf{5-shot} & \textbf{1-shot} & \textbf{5-shot}
        \\
        \hlineB{2.5}
        ProtoNet{$^{\dagger}$}\cite{protonet} & 62.90 & 84.13 & 47.37 & 68.96 
        \\ \hline
        ~+ SENet{$^{\dagger}$}\cite{senet} & 69.62 & 85.90 & 48.58 & 67.84 
        \\
        ~+ CBAM{$^{\dagger}$}\cite{cbam} & 69.21 & 85.37 & 48.10 & 70.03
        \\
        ~+ SA{$^{\dagger}$}\cite{selfattention} & 69.23 & 87.49 & 50.07 & 70.41
        \\
        ~+ TDM + IAM & \textbf{72.96} & \textbf{88.02} & \textbf{52.88} & \textbf{72.81}
        \\
        \hlineB{2.5}
    \end{tabular}
    \label{tab:general_attention_methods}
\end{table}
\endgroup



\ifCLASSOPTIONcompsoc
  \section*{Acknowledgments}
\else
  \section*{Acknowledgment}
\fi
This work was supported in part by MSIT\&KNPA/KIPoT (Police Lab 2.0, No. 210121M06) and MSIT/IITP (No. 2019-0-00421, 2020-0-01821).

\ifCLASSOPTIONcaptionsoff
  \newpage
\fi



%
\small
\bibliographystyle{ieee_fullname}
\bibliography{egbib}



%

\begin{IEEEbiography}
[{\includegraphics[width=1in,height=1.25in,clip,keepaspectratio]{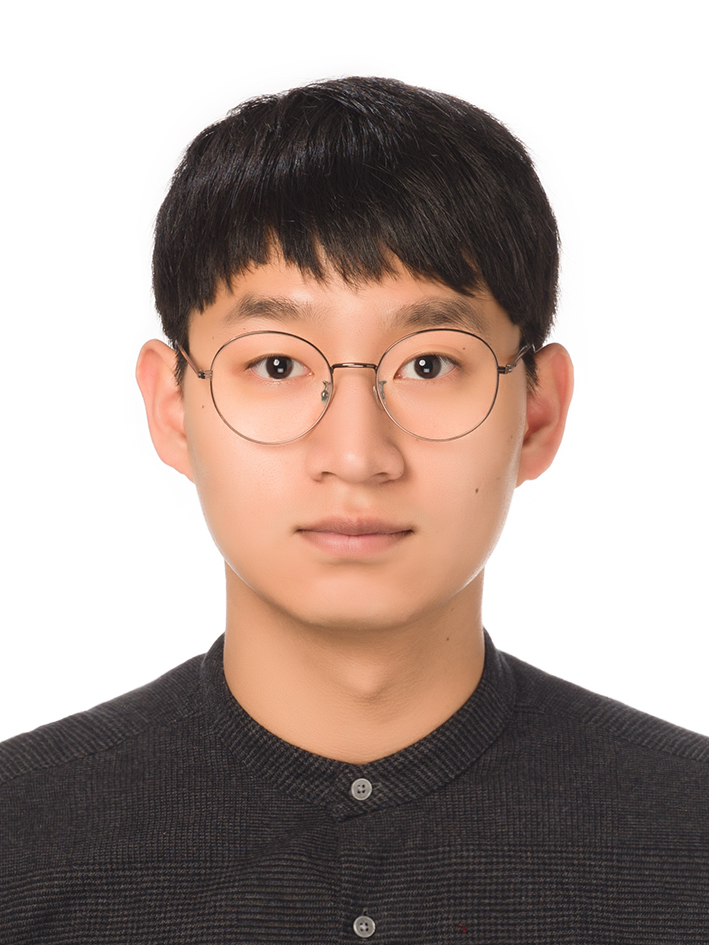}}]
{SuBeen Lee}
receives the B.S. degree in computer science from the Jeonbuk National University (JBNU), South Korea, in 2020, and the M.S. degree in artificial intelligence from Sungkyunkwan University (SKKU), South Korea, in 2022, where he is currently pursuing the Ph.D. degree in artificial intelligence.
His research interests include few-shot classification, attention, and deep learning.
\end{IEEEbiography}

\begin{IEEEbiography}
[{\includegraphics[width=1in,height=1.25in,clip,keepaspectratio]{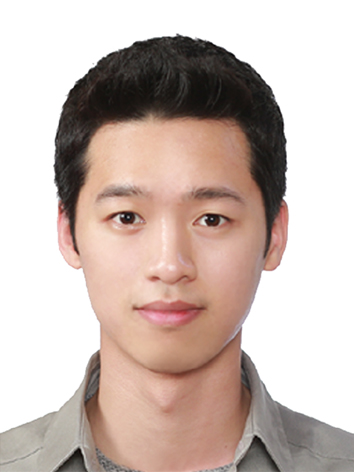}}]
{WonJun Moon} 
is a Ph.D. student at the Department of Artificial Intelligence, Sungkyunkwan University (SKKU), South Korea.
He received B.S., and M.S. degrees in computer science from Sungkyunkwan University (SKKU) in 2021 and 2022, respectively. Currently, his research areas include computer vision and deep learning.
\end{IEEEbiography}

\begin{IEEEbiography}
[{\includegraphics[width=1in,height=1.25in,clip,keepaspectratio]{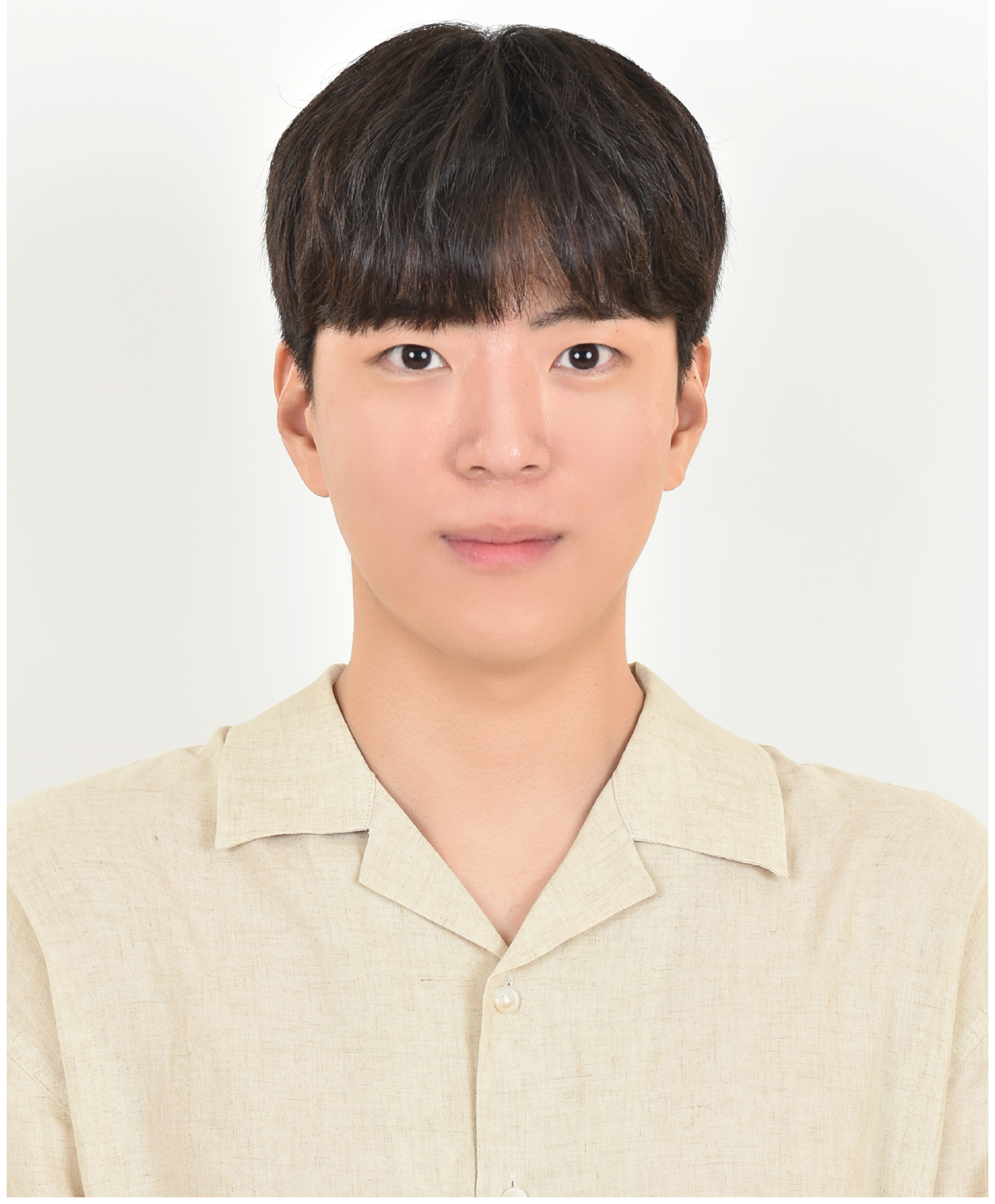}}]
{Hyun Seok Seong} 
received the B.S. degree in electronic and electrical engineering from Sungkyunkwan University (SKKU), South Korea, in 2019, where he is currently pursuing the combined M.S. and Ph.D. degrees in artificial intelligence. His research interests include metric learning for image categorization, machine learning, and deep learning.
\end{IEEEbiography}

\begin{IEEEbiography}
[{\includegraphics[width=1in,height=1.25in,clip,keepaspectratio]{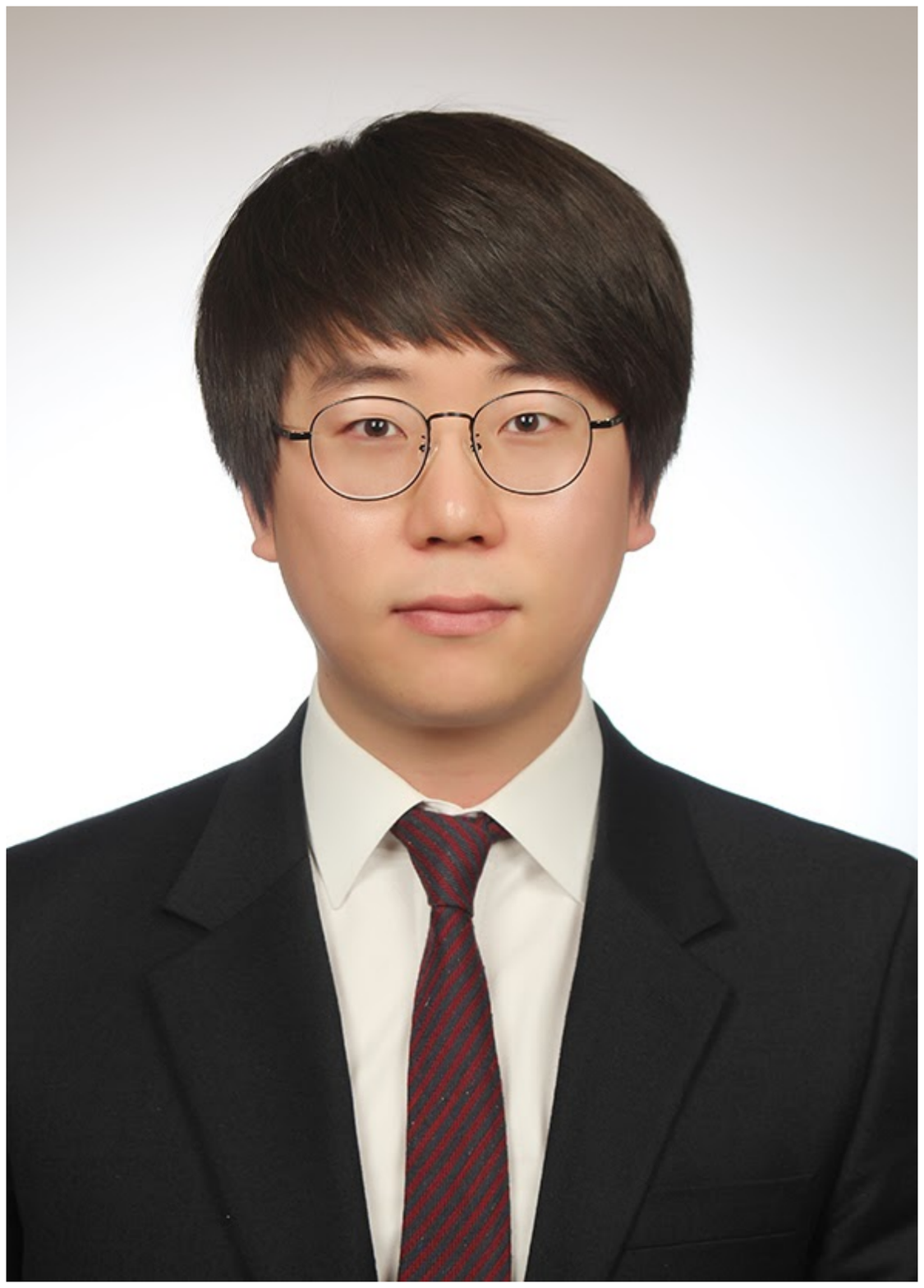}}]
{Jae-Pil Heo}
(Member, IEEE) received the B.S., M.S., and Ph.D. degrees in computer science from the Korea Advanced Institute of Science and Technology (KAIST), in 2008, 2010, and 2015, respectively. 
He is currently an Associate Professor at Sungkyunkwan University (SKKU), South Korea.
Before joining SKKU, he was a Researcher at the Electronics and Telecommunications Research Institute (ETRI). His research interests include computer vision and machine learning.
\end{IEEEbiography}







\end{document}